\documentclass[10pt,journal,compsoc]{IEEEtran}
\ifCLASSOPTIONcompsoc
  \usepackage[nocompress]{cite}
  \usepackage{amsmath,amsfonts}
  \usepackage{algorithmicx}
  \usepackage{algorithm}
  \usepackage{algpseudocode}
  \usepackage{array}
  \usepackage[caption=false,font=normalsize,labelfont=sf,textfont=sf]{subfig}
  \usepackage{textcomp}
  \usepackage{stfloats}
  \usepackage{url}
  \usepackage{verbatim}
  \usepackage{graphicx}
  \usepackage{cite}
  \usepackage{multirow}
\else
  \usepackage{cite}
\fi
\ifCLASSINFOpdf
\else
\fi
\begin{document}
%
\title{Dual Compensation Residual Networks for Class Imbalanced Learning}
%
%
%
%

\author{Ruibing~Hou,~\IEEEmembership{Student Member,~IEEE},
        Hong~Chang,~\IEEEmembership{Member,~IEEE}, Bingpeng~Ma,~\IEEEmembership{Member,~IEEE},
        Shiguang~Shan,~\IEEEmembership{Fellow,~IEEE}, 
        and~Xilin~Chen,~\IEEEmembership{Fellow,~IEEE}
\IEEEcompsocitemizethanks{\IEEEcompsocthanksitem R. Hou, H. Chang, S. Shan and X. Chen are with Key Laboratory of Intelligent Information Processing, Institute of Computing Technology (ICT), Chinese Academy of Sciences (CAS), Beijing, 100190, China, and University of the Chinese Academy of Sciences, Beijing 100049, China.
\IEEEcompsocthanksitem B. Ma is with the School of Computer Science and Technology, University of the Chinese Academy of Sciences, Beijing 100049, China. \protect\\
E-mail: ruibing.hou@vipl.ict.ac.cn, bpma@ucas.ac.cn, \{changhong, sgshan, xlchen\}@ict.ac.cn}
\thanks{Manuscript received April 19, 2005; revised August 26, 2015.}}

%
%

\markboth{Journal of \LaTeX\ Class Files,~Vol.~14, No.~8, August~2015}%
{Shell \MakeLowercase{\textit{et al.}}: Bare Demo of IEEEtran.cls for Computer Society Journals}
%



\IEEEtitleabstractindextext{%
\begin{abstract}
Learning generalizable representation and classifier for class-imbalanced data is challenging for data-driven deep models. Most studies attempt to re-balance the data distribution, which is prone to overfitting on tail classes and underfitting on head classes. In this work, we propose Dual Compensation Residual Networks to better fit both tail and head classes. Firstly, we propose dual \textit{\textbf{Feature Compensation Module}} (FCM) and \textit{\textbf{Logit Compensation Module}} (LCM) to alleviate the overfitting issue. The design of these two modules is based on the observation: an important factor causing overfitting is that there is severe \textit{feature drift} between training and test data on tail classes. In details, the test features of a tail category tend to drift towards feature cloud of multiple similar head categories. So FCM estimates a multi-mode feature drift direction for each tail category and compensate for it. Furthermore, LCM translates the deterministic feature drift vector estimated by FCM along intra-class variations, so as to cover a larger effective compensation space, thereby better fitting the test features. 
Secondly, we propose a \textbf{\textit{Residual Balanced Multi-Proxies Classifier}} (RBMC) to alleviate the under-fitting issue. Motivated by the observation that re-balancing strategy hinders the classifier from learning sufficient head knowledge and eventually causes underfitting, RBMC utilizes uniform learning with a \textbf{residual path} to facilitate classifier learning. Comprehensive experiments on Long-tailed and Class-Incremental benchmarks validate the efficacy of our method.
\end{abstract}

\begin{IEEEkeywords}
Class Imbalance Learning, Class-Incremental Learning, Residual Path
\end{IEEEkeywords}}

\maketitle

\IEEEdisplaynontitleabstractindextext

%
\IEEEpeerreviewmaketitle

\IEEEraisesectionheading{\section{Introduction}\label{sec:introduction}}
\section{Introduction}
Recently, many vision tasks have made significant progress with deep neural networks~\cite{residual}, driven by the development of deep neural networks as well as large-scale datasets~\cite{imageNEt}. However, these large-scale datasets are usually well-designed, and the number of samples in each class is balanced artifically. In real world, most data exhibits an \textit{extremely class imbalanced} (long-tail)  distribution~\cite{genome,Lvis}, where a few high-frequency (head) classes occupy most of the instances, while most low-frequency (tail) classes are under-represented by limited samples. The data imbalance poses great challenges to learning unbiased networks~\cite{Mateusz}. 

An intuitive solution to address class imbalance problem is to re-balance data distribution~\cite{khan2017cost,Balanced,cao2019learning,drummond2003c4}, including \textit{re-sampling} training data~\cite{drummond2003c4,Aditya,han2005borderline} and \textit{re-weighting} loss functions~\cite{re-weight,krawczyk2016learning,CBL}. 
However, the re-balanced distribution is easily fitted by the  over-parameterized deep networks, increasing the risk of under-fitting the whole original data distribution while over-fitting the tail data~\cite{Decoupling}.
Recent work~\cite{Decoupling} observes that uniform learning (\textsl{i.e.} training without re-balancing) actually learns more generalizable representation. Based on above observation,  the work~\cite{Decoupling} proposed a decoupled scheme. In general, the decoupled scheme first learns feature representation under \textbf{Uniform Sampling (US)}, and then re-adjusts the linear classifier on frozen features under \textbf{Class-Balanced Sampling (CBS)}. Nevertheless, such a two-stage decoupled scheme still tends to produce a biased classification boundary and fails to deal with imbalance issue well.

\begin{figure}[t]
\centering
\includegraphics[width=1.0\linewidth]{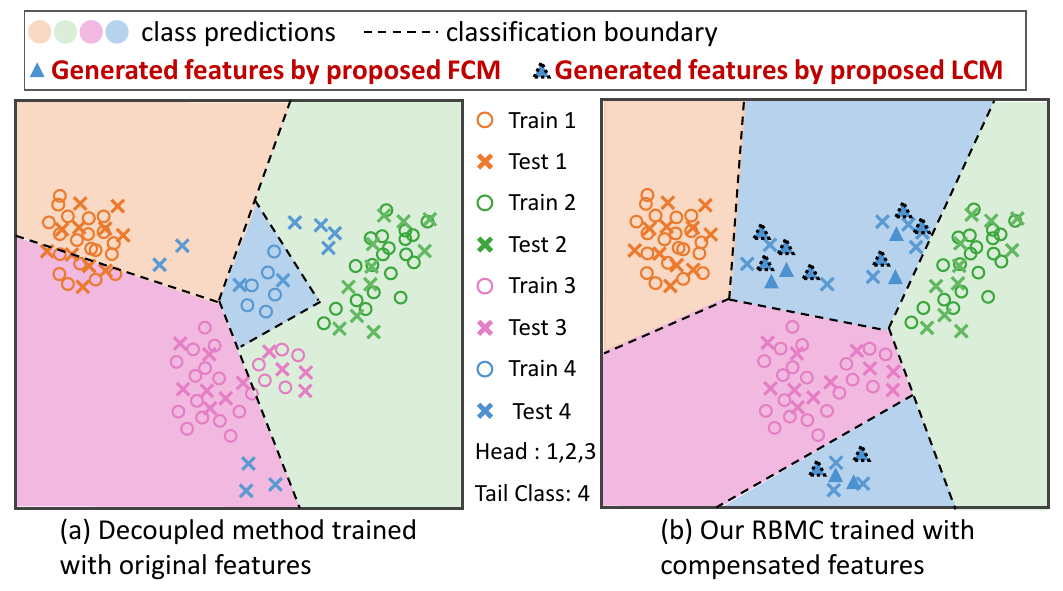}
\caption{The feature visualization with t-SNE~\cite{van2008visualizing}. (a) The classifier of decoupled method~\cite{Decoupling} \textit{over-fits} the biased training features on tail classes and \textit{under-fits} on head classes. (b) The proposed RBMC trained with compensated features by FCM and LCM could achieve a more accurate classification boundary, via better fitting both tail and head classes. }
\label{fig1}
\end{figure}

\begin{figure}[t]
\centering
\includegraphics[width=1.0\linewidth]{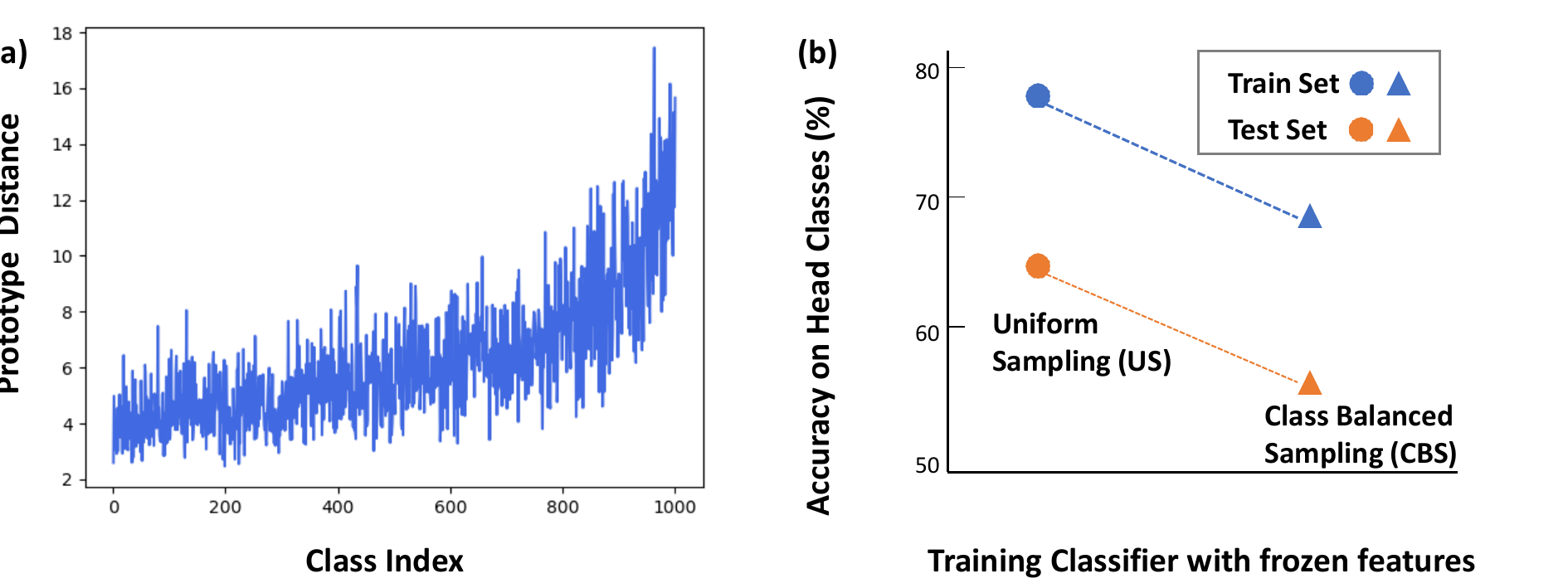}
\caption{Illustration of two problems of existing class-imbalanced methods \cite{Decoupling,BBN,CBL}. (a) The distance of per-class prototype between training and test data of existing method~\cite{Decoupling}. As seen, the distance goes large as the number of training samples decreases, indicting a \textit{large feature drift on tail classes}. (b) With frozen features, top-1 accuracy on head classes of classifier learning under US/CBS sampling \cite{Decoupling}. As observed, re-balanced classifier performs poorly on both training and test compared to uniform classifier, proving that re-balanced classifier essentially suffers from \textit{under-fitting the head classes}.}
\label{fig2}
\end{figure}

We argue that there are two main problems that make existing works \cite{Decoupling,BBN,khan2017cost,Balanced,cao2019learning,drummond2003c4,drummond2003c4,Aditya,han2005borderline,re-weight,krawczyk2016learning,CBL} produce \textit{biased classification boundary}. The first is the \textbf{feature drift between training and test data on tail classes}. For a tail class with scanty training samples, the feature extractor easily overly memorizes the knowledge of the few samples, which cannot generalize to unseen test samples~\cite{MAML}. So the extracted features of training and test samples  tend to deviate from each other. An example is illustrated in Fig.~\ref{fig1} (a) and Fig. ~\ref{fig2} (a)\footnote{The classes are re-indexed so that the number of training samples decreases as the class index increases}. As a consequence, the classifier trained on the drifted features would fail to separate the test samples of tail classes, as shown in Fig.~\ref{fig1} (a). The second is the \textbf{re-balanced classifier under-fitting on head classes}. In the phase of classifier learning, existing methods typically utilize the re-balancing strategy to obtain a class-balanced classifier. However, the re-balancing strategy encourages the classifier to excessively focus on tail classes, which inevitably leads to under-optimize on head classes. An example is illustrated in Fig.~\ref{fig2} (b). As a result, the classifier trained only by re-balancing strategy is insufficient to recognize all head samples, as shown in Fig.~\ref{fig1} (a). 

In this work, we propose \textbf{Dual Compensation Residual Networks}  (DCRNets) to learn a more unbiased classifier on imbalanced data. DCRNets contains three key modules: dual compensation component consisting of \textbf{Feature Compensation Module} (FCM) and \textbf{Logit Compensation Module} (LCM) to alleviate the overfitting on tail classes, and \textbf{Residual  Balanced Multi-Proxies Classifier} (RBMC) to alleviate the underfitting on head classes. The proposed  modules can serve as plug-and-play modules, seamlessly integrating with  existing methods with any network architecture, such as ResNet~\cite{residual}, ResNeXt~\cite{xie2017aggregated} or Multi-expert framework~\cite{wang2020long}, forming our DCRNets. 

 The first module, FCM, alleviates overfitting on  tail data by compensating for feature drift on tail classes. FCM is designed based on the following observation as shown in Fig.~\ref{fig1} : (1) The test tail features tend to drift towards the feature cloud of \textbf{similar head categories}. (2) The feature drift direction presents a multi-mode, where the test features of one tail category could drift towards \textbf{multiple} similar head categories. Based on above observations, FCM estimates a \textbf{multi-mode feature drift} direction for each tail category based on \textbf{nearest head categories}. Along the estimated feature drift vector, original tail features can be moved closer to the test features, as shown in Fig.~\ref{fig1} (b).

Although FCM can reduce the feature drift of tail classes, it  can be challenging to obtain entirely  accurate feature drift vectors due to the complex distribution often presented by test features of tail classes \cite{allen2019infinite}. To this end, we design the second module, LCM, which translates the deterministic feature drift direction estimated by FCM along intra-class variations. In this way, the compensated features can cover a larger feature space and still live in true feature manifold, thus can better fit the test features. Furthermore, instead of performing explicit translation,  LCM resorts to an efficient form of logits (classifier outputs) compensation that allows for end-to-end training. 
As shown in Fig.~\ref{fig1} (b), with FCM and LCM, the training features of a tail category can be compensated to corresponding test feature space, making the learned classification boundary more generalizable.

The last module, RBMC, alleviating underfitting on head data by introducing a residual path from uniform learning. Motivated by the fact that the classifier trained under uniform sampling (US) can better fit the head data~\cite{Decoupling}, RBMC is designed to utilize the  predictions of US classifier. In particular, RBMC learns a \textbf{residual} mapping from imbalanced predictions of US classifier to desired balanced predictions through class-balanced sampling (CBS) training. As a result, RBMC is  optimized jointly under both US and CBS strategies, so that it can take full advantage of all training data to alleviate under-fitting, while also avoiding highly skewing towards head classes.

For evaluation, we conduct Class Imbalanced experiments on four long-tailed benchmarks. We demonstrate that various long-tailed methods~\cite{Decoupling,wang2020long} can be directly incorporated  into the architecture of DCRNets, yielding consistent performance improvements. Further, we conduct experiments on a closely connected field, Class Incremental Learning, to further show the generality of our method.

\section{Related Work} 
\subsection{Long-tailed Classification}
Long-tailed classification is a long-standing research problem in machine learning, where the key is to overcome the severe class imbalance issue~\cite{shawe1999optimizing}. With the great success deep neural networks have achieved in balanced classification tasks, increasing attention is being shifted to long-tailed classification. Recent studies can be roughly divided into the following six categories. 
\\

\noindent
\textbf{Re-sampling/Re-weighting.} \quad
An intuitive solution to long-tailed task is to re-balance the data distribution. Re-sampling data is a common re-balancing strategy to artificially balance the imbalanced data. Typical re-sampling includes over-sampling~\cite{Relay,Mateusz,re-weight,zhong2019unequal}  by simply repeating data from tail classes, and under-sampling by discarding data from head classes~\cite{Mateusz,krawczyk2016learning,drummond2003c4}. However, duplicated tail data may cause over-fitting tail classes~\cite{chawla2002smote,CBL}. In addition, it is revealed that over-sampling leads to a side-effect of making head classes under-represented~\cite{Decoupling,BBN}. And discarding head data will certainly impair the generalization ability of deep networks~\cite{BBN}.

Re-weighting loss is another prominent re-balancing strategy. The basic idea of Re-weighting is to upweight the tail samples and downweight the head samples in the loss function~\cite{khan2017cost}.  The class-sensitive loss~\cite{huang2016learning,khan2017cost,huang2019deep} assigned the weight to each class inversely proportional to the number of samples. Cui \textsl{et al.}~\cite{CBL} proposed to adapt the effective number of samples instead of proportional frequency. A more fine-grained usage is at sample level,  \textit{e.g.} focal loss~\cite{lin2017focal}, using meta learning~\cite{shu2019meta,MetaReWeight} or Bayesian uncertainty~\cite{khan2019striking}. The work~\cite{cao2019learning} designed a label-distribution-aware loss to encourage a larger margin for the tail class. However, It is revealed that re-weighting is not capable to handle large-scale long-tailed data and easily causes optimization difficulty~\cite{Decoupling,BBN}.
\\

\noindent
\textbf{Decoupled Methods.} \quad
Kang \textsl{et al.}~\cite{Decoupling} firstly proposed the decoupled scheme. This work observed that an uniform sampling scheme actually generated more generalizable representations and achieved stronger performance after re-balancing the classifier. So \cite{Decoupling}  proposed a two-stage decoupled training pipeline that  learned features using US at the first stage, then  re-adjusted the classifier with frozen features using CBS at the second stage. However, such a two-stage design is not for end-to-end frameworks. The works~\cite{tang2020long,menon2020long} improved the two-stage strategy by introducing a post-process to adjust the prediction score. Zhou \textsl{et al.}~\cite{BBN} simulated the two-stage training in  a single stage by dynamically combining uniform sampling and class balanced sampling with cumulative learning strategy. 

In spite of excellent results, above decoupled methods do not take into account the feature drift of tail categories, leading to a biased learning of classifier.  On the contrary, our FCM and LCM effectively reduce the unfavorable feature drift. In addition, the decoupled methods typically only rely on class balanced sampling to optimize the classifier. So  the classifier learning inevitably suffers from adverse effects of class balanced sampling, \textit{i.e.}, a higher risk of under-fitting on head data.  On the contrary, our RBMC utilizes the classifier trained with uniform sampling, which can take full advantage of all training data, thereby alleviating the under-fitting on head classes.
\\

\noindent
\textbf{Self-supervised Contrastive Learning.} \quad
Self-supervised contrastive learning \cite{he2019moco,chen2020mocov2} trains the model in a pairwise way by aligning positive sample pairs and repulsing negative sample pairs. Recently, researchers have applied self-supervised learning to long-tailed recognition and demonstrated improved performance. SSP~\cite{SSL} applied self-supervised learning~\cite{he2019moco} for a good feature initialization. KCL~\cite{kang2021exploring} further designed a balanced contrastive loss. However, above works~\cite{SSL,kang2021exploring} require a two-stage learning paradigm, which are not for end-to-end frameworks. Hybrid-SC \cite{wang2021contrastive} proposed a one-stage framework that used a supervised contrastive  loss to learn better representation and a cross-entropy classification loss to learn a balanced classifier at the same time. Furthermore, PaCo \cite{cui2021parametric} and GPaCo \cite{cui2022generalized} applied re-balance strategy in supervised contrastive learning to tackle imbalance issue. Notably, self-supervised contrastive learning techniques are orthogonal to our method, which can be easily combined to learn better representation and further improve performance.
\\

\noindent
\textbf{Transfer Learning.} \quad
Another line is to transfer knowledge learned from head classes to tail classes.   Some works~\cite{imageNEt,zhu2020inflated} exploited complex memory backs to transfer semantic features.  Wang~\textsl{et al.}~\cite{wang2017learning} designed a meta network to transfer the meta-knowledge of parameters evolution. Kim~\textsl{et al.}~\cite{kim2020m2m} translated the head samples to a synthetic tail class via a complicated optimizing process.  Chu~\textsl{et al.}~\cite{chu2020feature} generated  tail features by mixing its class-specific features with the class-generic features transferred from head classes.  Nevertheless, above methods required either large memory of historical features~\cite{imageNEt,zhu2020inflated} or complex optimization process~\cite{wang2017learning,kim2020m2m,chu2020feature}, incurring high cost in time and memory. Recent methods~\cite{liu2020deep,FTL,hariharan2017low} proposed to transfer the intra-class variance from head class to enlarge the diversity of tail classes. However, ~\cite{liu2020deep,FTL,hariharan2017low} made strong assumptions that each class has a shared variance, which could introduce harmful features that interfered with tail features~\cite{zang2021fasa}. And recent works~\cite{yang2021free,yang2021delving} proposed to transfer the mean and variance statistics for few-shot classification and long-tailed regression. But \cite{yang2021free} focused on few-shot task, which could not distinguish the tail from head classes. And \cite{yang2021delving} aimed to create a continuity in feature space of continuous labels for regression task, which could not recognize discrete classes for recognition task.  

Our FCM shares a similar idea that the head categories can be used to help the learning of tail categories. But to the best of our knowledge, our FCM is the first to point out the test features of tail category tend to drift towards the feature space of similar head categories. Therefore, FCM proposes to compensate the feature drift of tail categories with the information of similar head categories. In implementation, FCM is simple and waives the need for complex memory and optimization mechanism. So FCM is more easier to combine with other advanced long-tailed learning frameworks, such as multi-expert network~\cite{wang2020long}.
\\

\noindent
\textbf{Data Generation.} \quad
In the context of long-tailed recognition, the tail class essentially lacks of sufficient diversity to learn a generalizable model. Therefore, one remedy is to generate more tail samples. In this part, we focus on the approaches that \textbf{performed generation without using head information}. 
For example, Zhang~\textsl{et al.}~\cite{zang2021fasa} estimated a class-wise feature distribution  with statistics calculated from observed samples, and then generated virtual features for tail classes based on the estimated distribution. Wang~\textsl{et al.}~\cite{wang2018low} used training samples from tail classes and noise vectors to produce new hallucinated tail samples. Mullick~\textsl{et al.}~\cite{mullick2019generative}  used the convex combination of existing samples to generate new tail samples. Recent works~\cite{li2021metasaug,li2021transferable} used Implicit Semantic Data Augmentation (ISDA) algorithm~\cite{wang2019implicit} to perform implicit semantic augmentation. 

Our LCM shares a similar idea of generating more diverse tail samples. However, different from most works~\cite{zang2021fasa,wang2018low,mullick2019generative} that explicitly generate samples, LCM performs an implicit generation which is highly efficient and easy to combine with other advanced long-tailed methods. Notably, ISDA~\cite{wang2019implicit} also implicitly generates samples. But ISDA generates new features around the training instances of the target class. In long-tail classification task, due to the feature drift of tail classes, the generated features would still be far away from the real test features. Thus ISDA has limited effectiveness in learning a generalized model for long-tailed data. Differently, LCM utilizes an extra feature drift compensation operation, which can generate more realistic features to improve the generalizability of long-tailed models. 
\\

\begin{figure*}[t]
\centering
\includegraphics[width=0.75\linewidth]{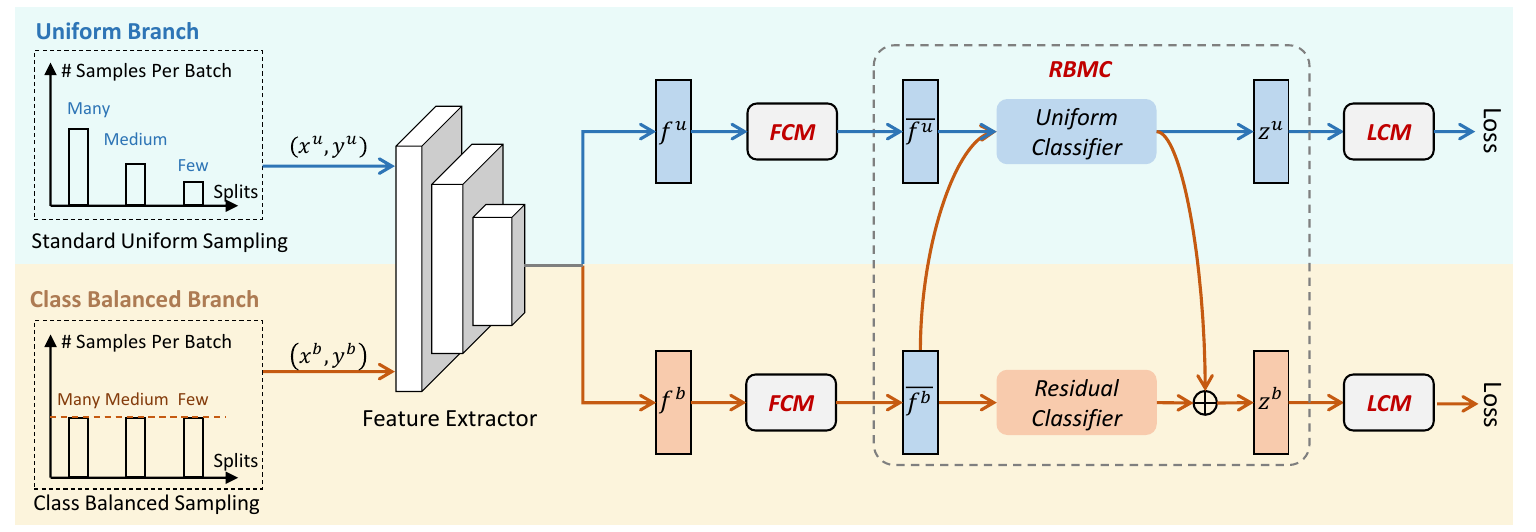}
\caption{Framework of Dual Compensation Residual Networks. It contains three key parts: \textit{Feature Compensation Module} (FCM) to compensate the feature drift of tail categories, \textit{Logit Compensation Module} (LCM) to implicitly enrich feature diversity, and \textit{Residual Balanced Multi-proxies  Classifier} (RBMC) to learn a class-balanced classification boundary.}
\label{fig3}
\end{figure*}

\noindent
\textbf{Multi-Expert Networks.}
The recent trend  in multi-expert networks shows their strong potential to effectively address long-tailed issue. LFME~\cite{xiang2020learning-19} and RIDE~\cite{wang2020long} are two typical multi-expert architectures.  The two methods learned diverse models in parallel, combining with knowledge distillation and distribution-aware expert selection. Further, ACE~\cite{cai2021ace} proposed to use complementary experts, where each expert is most knowledgeable in a specific subset and complementary to other experts.  Similarity, ResLT~\cite{cui2022reslt} divided classes into several subsets, \textit{i.e.},  all classes, middle and tail classes and tail classes. ResLT then trained multiple experts with various classes subsets under a residual learning mechanism. Notably, ResLT shares most parameters for different experts, which has less computational cost than other multi-expert networks~\cite{xiang2020learning-19,wang2020long,wang2020long}.
NCL~\cite{li2022nested} collaboratively learned multiple experts concurrently and used knowledge distillation strategy~\cite{zhang2018deep} to enhance each single expert. TADE~\cite{zhang2021test} extended the multi-expert network to test-agnostic long-tailed recognition task, where the training class distribution is long-tailed while the test class distribution is unknown and can be skewed arbitrarily. With the ensemble methods, the multi-expert networks set new state-of-the-art performance on standard long-tailed benchmarks. We validate in the experiments that our approach can be effectively inserted into the multi-expert networks to bring consistent performance improvements.

\subsection{Class Incremental Learning}
Incremental learning \cite{delange2021continual} aims to learn efficient models from the data that gradually comes in a sequence of training phases. Class Incremental Learning (CIncL) \cite{rebuffi2017icarl} is a thriving subfield in incremental learning, where each phase has the data of a new set of classes coming from the identical dataset. Related methods mainly focus on how to solve the problems of forgetting old data, which can be roughly categorized into regularized-based and replay-based methods.  Regularized-based methods \cite{li2017learning,douillard2020podnet,tao2020topology} introduce regularization terms in the loss function to consolidate previous knowledge when learning from new data. Replay-based methods \cite{rebuffi2017icarl,liu2020mnemonics} use the rehearsal strategy, which store a tiny subset of old data, and replay the model on them to reduce the forgetting.  

For class incremental learning, data of old classes is generally not available when new classes appear. Even with the rehearsal strategy, the ratio of the number of new samples to that of old exemplars could be very high,  leading to a very serious imbalance issue in class incremental learning \cite{he2021tale}. In order to address the class imbalance issue in class incremental learning, Castro \textsl{et al.} \cite{castro2018end} utilized a balanced fine tuning strategy. Belouadah \textsl {et al.} \cite{belouadah2019il2m} stored the average confidence at each incremental phase to rectify the imbalanced logits. Wu \textsl{et al.} \cite{wu2019large} added a bias correction layer to correct the output logits. Hou  \textsl{et al.} \cite{hou2019learning} combined cosine normalization, less-forget constraint and inter-class separation to mitigate the impact of class imbalance. Zhao \textsl{et al.}  \cite{zhao2020maintaining} proposed weight aligning to correct the biased weight in classification layer after uniform training process, which is similar to the $\tau$-norm classifier of decoupled method \cite{Decoupling}.  

Above methods show that the techniques to address class imbalance can be applied to Class Incremental Learning. From the class imbalance view, most above methods \cite{wu2019large,hou2019learning,zhao2020maintaining} combine a uniform training  stage on the imbalanced dataset and a balanced fine-tuning stage, similar to the two-phase decoupled strategy \cite{Decoupling}. However, these methods neglect the issue of feature drift, which causes overfitting on the few stored old exemplars and poor performance on old classes. Our FCM and LCM can effectively alleviate the unfavorable feature drift, and improve classification performance under class imbalance, thereby benefiting class incremental learning.

\section{Dual Compensation Residual Networks}
In this section, we first present the overall framework of DCRNets. Then, we elaborate on the dual compensation modules and the residual balanced classifier used as  parts of DCRNets. We finally give the detailed training and test algorithm of DCRNets.

\subsection{Overall Framework}
Fig.~\ref{fig3} shows the overview of proposed DCRNets. The network consists of two branches: one Uniform Branch equipped with Uniform Sampling (US) and one Class Balanced Branch equipped with Class Balanced Sampling (CBS). A feature extractor (backbone) is shared between the two branches, and each branch has its own classifier. The Uniform Branch mainly aims to learn generalizable features. And the Class Balanced Branch is expected to learn a less biased classifier. Building on the two-branch architecture, DCRNets contains three novel parts: FCM, LCM and RBMC. FCM and LCM aims to alleviate overfitting on tail classes by compensating the feature drift between training and test data. RBMC aims to alleviate underfitting on head classes by adding a residual path from uniform branch to class balanced branch.   

Formally, given an input image $\mathbf{x}^u/\mathbf{x}^b$ sampled under US/CBS, the shared backbone is firstly used to extract the feature representation $\boldsymbol{f}^u/\boldsymbol{f}^b$ for Uniform Branch/Class Balanced Branch. Motivated by the observation that there is severe feature drift between training and test data on tail classes, the original feature vectors ($\boldsymbol{f}^u$ and $\boldsymbol{f}^b$) are sent into FCM to compensate the drift, obtaining compensated features $\overline{\boldsymbol{f}^u}$ and $\overline{\boldsymbol{f}^b}$. After that, the compensated feature $\overline{\boldsymbol{f}^u}$ is sent into the uniform classifier of Uniform Branch for feature learning, and $\overline{\boldsymbol{f}^b}$ is sent into the RBMC of Class Balanced Branch for classifier learning. To  prevent the classifier from fitting to FCM errors and enrich feature diversity especially for tail classes, 
the outputs of uniform classifier and RBMC are adjusted by LCM. At the last, DCRNets are optimized by applying cross-entropy classification loss on top of LCM.

\subsection{Feature Compensation Module}
\label{sec-fcm}

In this section, we first analyze the feature drift phenomenon in detail, and then present the specific operations of FCM. 

\begin{figure}[t]
\centering
\includegraphics[width=0.75\linewidth]{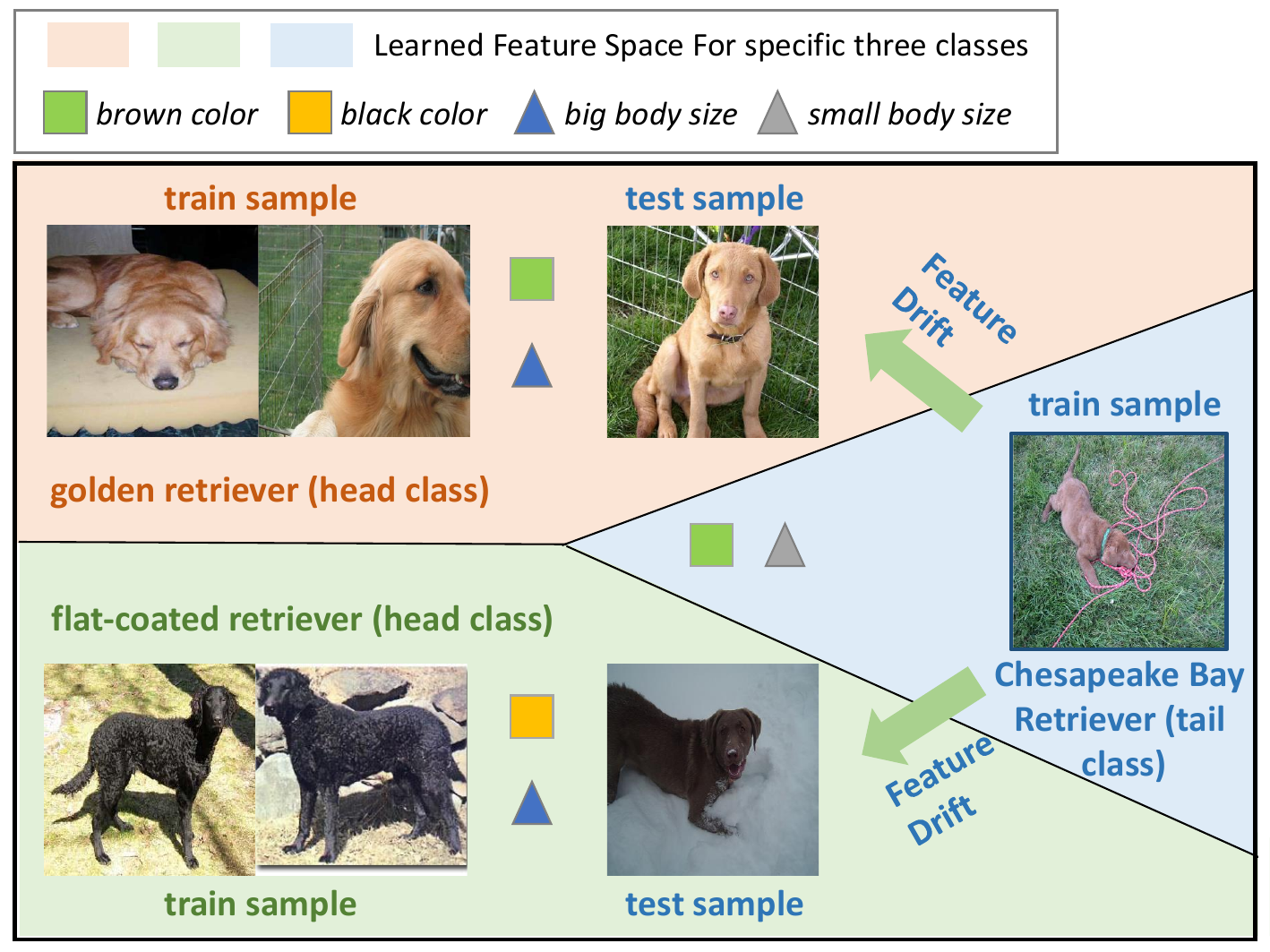}
\caption{Due to some common characteristics, the features of two test samples of tail category ``Chesapeake Bay retriever'' would drift towards the feature space of similar head categories ``golden retriever'' and ``flat-coated retriever'', respectively.}
\label{fig4}
\end{figure}

\noindent \textbf{Analysis of Feature Drift.} \quad
For imbalanced data, the tail categories typically cannot provide sufficient diversity to learn generalizable features. So there could be discrepancy in feature space between training and test data on tail categories. Fig.~\ref{fig2} (a) computes the Euclidean Distance of per-class feature prototype between training and test data. As seen, the distance  increases as the number of training samples decreases, indicating a larger feature drift on tail categories. As a result, the classifier learned to differentiate the training samples can not generalize to differentiate test samples for tail classes. 

\begin{figure}[t]
\centering
\includegraphics[width=1.0\linewidth]{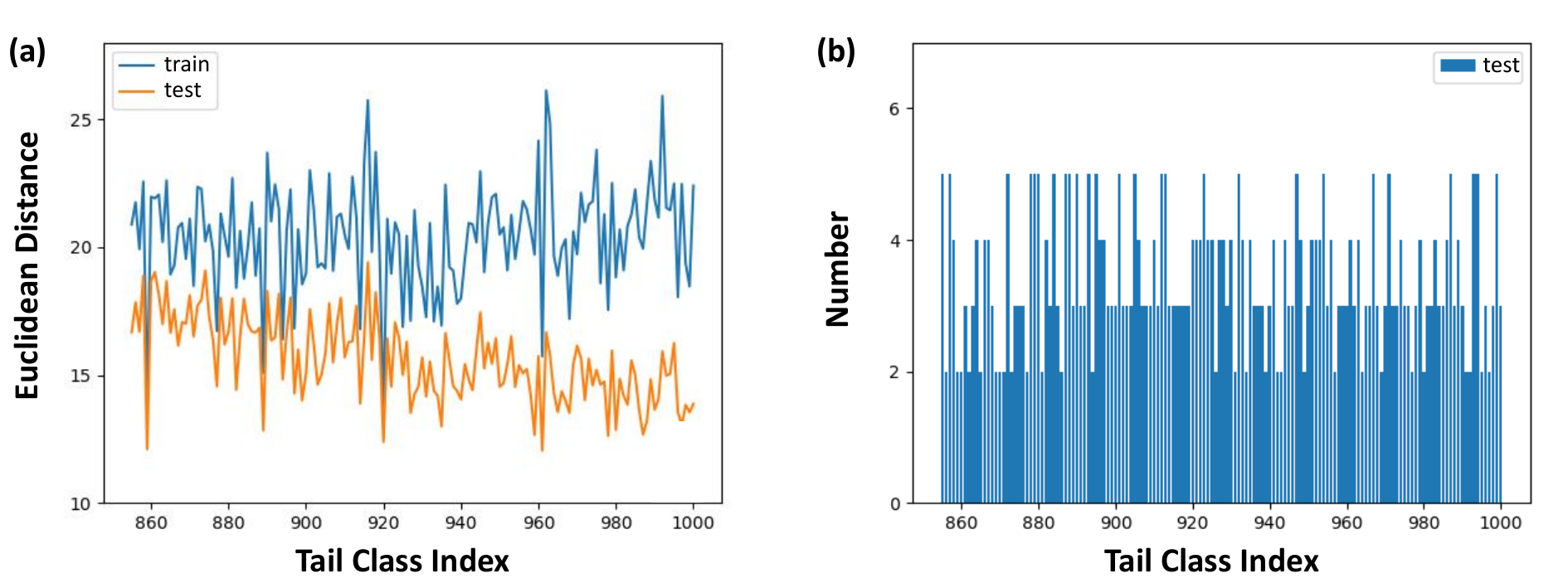}
\caption{(a) For each tail category on train/test set, the average distance of all of its features to the prototype of nearest head category.  The tail features of \textbf{test set} become \textbf{closer} to the nearest head category, indicating a drift towards similar head features. (b) For each tail category on test set, the number of different nearest head categories across all samples. On test set,  the samples of one tail category could be closest to different head categories. The result indicates that the tail features tend to drift towards \textbf{multiple} similar head categories.}
\label{fig5}
\end{figure}

Furthermore, we analyze the direction of feature drift.  \textbf{For one thing, we observe that the test features of tail category tend to drift towards the feature clouds of similar head categories}. In detail, since tail category contains scanty training samples, the feature extractor typically learns \textbf{incomplete} features that over-fit  to the scanty training samples. For example, as shown in Fig.~\ref{fig4}, the  \textit{brown color} and \textit{small size} features are enough to recognize the training samples of \textit{Chesapeake Bay retriever} tail category. But the two features are insufficient to correctly recognize the unseen test samples of \textit{Chesapeake Bay retriever}. Meanwhile, because test samples of tail category often share some characteristics with similar head categories, the test features of tail category tend to drift towards the feature space of similar head categories, as illustrated in Fig.~\ref{fig4}. Fig.~\ref{fig5} (a) calculates the average distance between all train/test features of each tail category and the prototype of the nearest head category. As seen, as the number of samples in the tail category decreases, the test features of tail category become closer to the nearest head category. The result verifies that the test features of tail category could drift  towards feature space of similar head categories.  

\textbf{For another, we observe that the direction of feature drift presents a multi-mode.} In details, the test samples of one tail category usually have various characteristics, \textsl{e.g.}, various shapes, sizes and poses. Thus, different test samples could share characteristics with different head categories, resulting in a drift towards multiple similar head categories. For example, in Fig.~\ref{fig4}, the top test sample has the same \textit{brown color} and \textit{big body} as \textit{golden retriever} category, while the bottom test sample has the same \textit{black color} as \textit{flat-coated retriever} category. As a result, the features of the two test samples drift towards \textit{golden retriever} and \textit{flat-coated retriever} respectively. 
Fig.~\ref{fig5} (b)  counts the number of different nearest head categories across all test samples for each tail category. We observe that most tail categories have more than one nearest head category. The result indicates that the tail features tend to drift towards the feature space of multiple head categories.
\\

\begin{figure*}[t]
\centering
\includegraphics[width=0.9\linewidth]{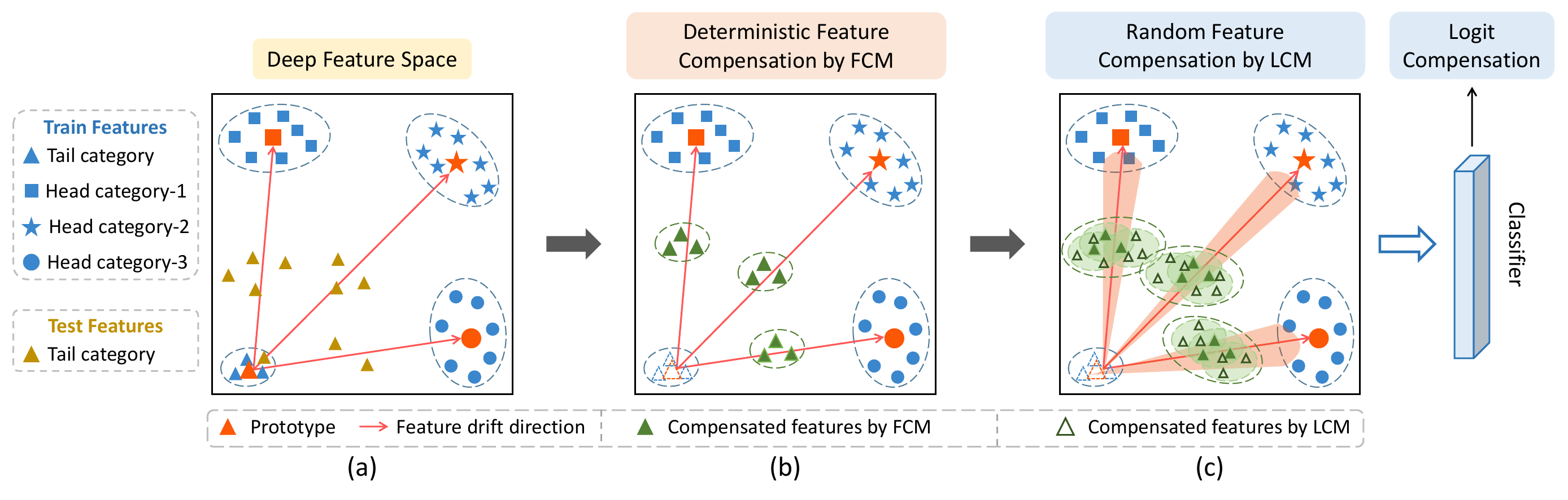}
\centering
\caption{Illustration of FCM and LCM. For each tail category (triangle), \textbf{(a)} FCM first computes multiple feature drift vectors (red lines) based on the distance vectors between the tail-category prototype and multiple similar head-category prototypes. \textbf{(b)} Then FCM shifts the original features (blue triangles) along the calculated drift vectors to obtain compensated features (green solid triangles). \textbf{(c)} LCM integrates uncertainty (red fan-shaped areas) into the deterministic drift vectors estimated by FCM, and generates richer  compensated features (green hollow triangles). }
\label{fig6}
\end{figure*}

\noindent \textbf{Computing of Feature Drift.} \quad
Aiming at reducing feature drift, we propose FCM to compensate for the training features. The key idea of FCM is to estimate a \textit{multi-mode feature drift direction} for each tail category based on the class similarity. 

Formally, FCM first computes a \textit{prototype} $\boldsymbol{c}_k \in \mathbb{R}^{D}$ for each category $k$. The prototype is defined as the mean vector of the training features belonging to its corresponding class \cite{prototypical}. Then for each tail category $t$, we select top $m$ head categories with the closest distance to the prototype of this tail category:
\begin{equation}
\begin{split}
&g\left(\boldsymbol{c}_t, \boldsymbol{c}_j\right) = \frac{\boldsymbol{c}_t^T\boldsymbol{c}_j}{\left\| \boldsymbol{c}_t\right\| \left\| \boldsymbol{c}_j\right\|}, \quad t \in C_t, \\
&\mathbb{D}_t = \{ g\left(\boldsymbol{c}_t, \boldsymbol{c}_j\right) \ | \ j \in C_h \}, \\
&\mathbb{S}_t = \{ j \ | \ g\left(\boldsymbol{c}_t, \boldsymbol{c}_j\right)  \in \rm{top_m}\left(\mathbb{D}_t\right) \}, \\
\end{split}
\label{eq1}
\end{equation}where 
$C_h$ and $C_t$ denote the set of head and tail categories respectively, 
$\rm{top_m}(\cdot)$ is an operator to select the top $m$ elements from the input distance set $\mathbb{D}_t$. 
$\mathbb{S}_t$ stores $m$ nearest head categories with respect to the tail category $t$.  

As above analysis, the test features of one tail category could drift towards multiple similar head categories. Therefore, we estimate a set of feature drift vectors for each tail category $t$ based on the nearest head categories in $\mathbb{S}_t$. As shown in Fig.~\ref{fig6} (a), since the prototype refers to class representative points in feature space, the prototype distance of tail category to the similar head category in $\mathbb{S}_t$ can be used to estimate corresponding feature drift vector:
\begin{equation}
\boldsymbol{\delta}_{tj} = \alpha_t \left(\boldsymbol{c}_j - \boldsymbol{c}_t\right), \quad j \in \mathbb{S}_t,
\label{eq2}
\end{equation}where  
$\alpha_t$ is a positive coefficient to control the strength of drift compensation. As the generalization of learned features is often positively correlated with the number of training samples, the category with fewer training samples is more likely to be closer to similar head category, as shown in  Fig.~\ref{fig5} (a). 
So $\alpha_t$ is set inversely proportional to the number of training samples of the tail category $t$. It is defined as:
\begin{equation}
\alpha_t = \alpha^0 \left(N_\text{max}-N_t\right)/\left(N_\text{max}-N_\text{min}\right),
\label{eq-a}
\end{equation}where 
$\alpha^0$ is a hyperparameter, and $N_\text{max}$ and $N_{\text{min}}$ are the numbers of training samples for the most frequent and least frequent category in $C_{t}$ respectively. 
\\ 

 \noindent 
\textbf{Compensation of Feature Drift.} \quad
Based on the estimated feature drift vectors  (Eq.~\ref{eq2}), FCM compensates for original training features to reduce the feature drift. In details, for each tail category $t$, we compute the probability $s_{tj}$ of drifting towards similar head category $j$ in $\mathbb{S}_t$, and the probability $s_{tt}$ of retaining its feature:
\begin{small}
\begin{align}
s_{tj} &= \frac{\exp\left(\tau g\left(\boldsymbol{c}_t, \boldsymbol{c}_j\right)\right)}{\sum_{l\in \mathbb{S}_t} \exp\left(\tau g\left(\boldsymbol{c}_t, \boldsymbol{c}_l\right)\right) + \exp\left(\tau g\left(\boldsymbol{c}_t, \boldsymbol{c}_t\right)\right)}, \quad j \in \mathbb{S}_t, \label{eq3} \\
s_{tt} &= \frac{\exp\left(\tau g\left(\boldsymbol{c}_t, \boldsymbol{c}_t\right)\right)}{\sum_{l\in \mathbb{S}_t} \exp\left(\tau g\left(\boldsymbol{c}_t, \boldsymbol{c}_l\right)\right) + \exp\left(\tau g\left(\boldsymbol{c}_t, \boldsymbol{c}_t\right)\right)},   
\label{eq4}
\end{align}
\end{small}where
$\tau$ is the temperature hyperparameter. 
Eq.~\ref{eq3} shows that the tail category has a higher probability of drifting towards more similar head category. 
Eq.~\ref{eq4} shows that the tail category is more likely to retain its features if it has a overall low similarity to all head categories in $\mathbb{S}_t$. 
This operation prevents the tail category from updating to a unrelated head category if this tail category  has no similar characteristics to all head categories. 

At last, for each training sample $\left(\boldsymbol{x}, t\right)$\footnote{For simplicity, we omit superscript $u$ and $b$ of all symbols in Section \ref{sec-fcm} and \ref{Section3.2}. } of tail category $t$, FCM shifts original feature $\boldsymbol{f}\in\mathbb{R}^{D}$ after backbone along the estimated drift vector $\boldsymbol{\delta}_{tj}$ with probability $s_{tj}$, to obtain a compensated feature $\overline{\boldsymbol{f}}\in\mathbb{R}^{D}$. FCM is formulated as: 
\begin{equation}
\overline{\boldsymbol{f}}=\boldsymbol{f}+\boldsymbol{r}, \ \text{where} \ p(\boldsymbol{r} = \boldsymbol{\delta}_{tj}) = s_{tj}, \ j\in \mathbb{S}_t \cup \{t\}.
\label{eq5}
\end{equation}
Eq.~\ref{eq5} is  equivalent to:
\begin{equation}
\overline{\boldsymbol{f}_j}=\boldsymbol{f}+\boldsymbol{\delta}_{tj}, \ \text{where} \ p(\overline{\boldsymbol{f}_j}) = s_{tj}, \ j\in \mathbb{S}_t \cup \{t\},
\label{eq6}
\end{equation}where
$\boldsymbol{\delta}_{tt}=\boldsymbol{0}$, indicating retaining original feature. 
Notably, we maintain training features of head categories, which are already fairly close to test features. As shown in Fig.~\ref{fig1} (b), when applying feature drift vectors to corresponding training features, the training features can be moved to be close to corresponding test space. In this way, the compensated features by FCM can better represent the test feature distribution,  thereby achieving a more generalizable classifier.

Notably, although FCM uses prototypes, it does not employ contrastive learning. The prototypes on FCM are directly computed and unlearnable, which are only used to estimate the feature drift direction $\boldsymbol{\delta_{tj}}$ (Eq.~\ref{eq2}) for tail classes. With the updated features $\overline{\boldsymbol{f}_j}$ (Eq.~\ref{eq6}) by FCM, we only employ \textit{classification} learning with a cross-entropy classification loss.

\subsection{Logit Compensation Module}
\label{Section3.2}
While FCM can roughly estimate the feature drift direction,  it is important to note that the test features of a tail class are unknown during training and typically exhibit complex  distribution. As a result, the drift vectors estimated by FCM  are prone to errors and difficult to cover all drift directions. To this end, we propose LCM which integrates uncertainty into drift estimation. Integrating uncertainty can prevent the subsequent classifier from over-fitting  to FCM errors, thereby enhancing its robustness. Also, integrating uncertainty can expand estimated region to cover more drift directions, so that more diverse features for tail classes can be generated to improve the classification boundary. 

As shown in Fig.~\ref{fig6} (c), LCM integrates uncertainty into the  \textbf{deterministic} drift vectors estimated by FCM ($\boldsymbol{\delta}_{tj}$ in Eq.~\ref{eq2}), via translating $\boldsymbol{\delta}_{tj}$ along intra-class variations.  Concretely, we establish a zero-mean Gaussian Distribution $\mathcal N \left(0, \boldsymbol{\sigma}_{t}^2 \right)$ for each category $t$, where $\boldsymbol{\sigma}_{t}\in\mathbb{R}^{D}$ is the class-conditional \textit{sample standard deviation} estimated from all training features of category $t$. \footnote{We assume the dimensions of the feature vector are independent of each other and consider the covariance matrix of the feature distribution to be diagonal. It can largely reduce the computation complexity and memory overload.}
Then a \textbf{random} drift vector $\overline{\boldsymbol{\delta}_{tj}}$ is obtained by translating $\boldsymbol{\delta}_{tj}$ along a random direction sampled from $\mathcal N \left( \boldsymbol{0}, \boldsymbol{\sigma}_{t}^2 \right)$. It is formulated as: 
\begin{equation}
\overline{\boldsymbol{\delta}_{tj}} = \boldsymbol{\delta}_{tj} + \beta_{t} \boldsymbol{g}, \ \text{where} \   \boldsymbol{g} \sim \mathcal N \left( \boldsymbol{0}, \boldsymbol{\sigma}_{t}^2 \right), \ j\in \mathbb{S}_t \cup \{t\},
\label{eq7}
\end{equation}where
$\beta_t$ is a positive coefficient to control the strength of translating $\boldsymbol{\delta}_{tj}$. Equivalently, we have $\overline{\boldsymbol{\delta}_{tj}}  \sim \mathcal  N \left(\boldsymbol{\delta}_{tj}, \beta_{t}\boldsymbol{\sigma}_{t}^2 \right)$. Based on $\overline{\boldsymbol{\delta}_{tj}}$, the compensated feature $\overline{\boldsymbol{f}}$ can be obtained by applying it to original  feature $\boldsymbol{f}$ with probability $s_{tj}$, 
\begin{equation}
\overline{\boldsymbol{f}} =\boldsymbol{f}+\boldsymbol{v}, \ \text{where} \ \boldsymbol{v}  \sim \sum_{j\in \mathbb{S}_t \cup \{t\}} s_{tj} \mathcal  N \left(\boldsymbol{\delta}_{tj}, \beta_{t}\boldsymbol{\sigma}_{t}^2 \right),
\label{eq8}
\end{equation}where 
$\boldsymbol{v}$ is the translation of the  original feature $\boldsymbol{f}$, which is  sampled from a \textit{Gaussian Mixture Distribution}. With probability $s_{tj}$, the compensated feature will shift towards the $j^{th}$ similar head class ($+\boldsymbol{\delta}_{tj}$) and vary along intra-class variance ($\beta_{t}\boldsymbol{\sigma}_{t}^2$). In this way, more diverse features that live in test manifold can be obtained, thereby achieving a more generalizable classifier. Notably, FCM (Eq.~\ref{eq5}) is a special case of Eq.~\ref{eq8} where the standard deviation of Gaussian Distribution is equal to $\boldsymbol{0}$ ($\boldsymbol{\sigma}^2_t=\boldsymbol{0}$).
\\ 

\noindent
\textbf{Class Adaptive Translation Coefficient.} \quad
Previous works~\cite{liu2020deep}  observe that due to lack of intra-class diversity, the class with fewer training instances usually has a smaller \textit{observed} variance. So the value of \textit{observed standard deviation} ($\boldsymbol{\sigma}^2_t$) for rarer class is relatively smaller,  which harms the diversity of generated features by LCM (Eq.~\ref{eq8}). 
To address this issue, existing works~\cite{liu2020deep,FTL} rely on a shared variance for all classes. But the strong assumption on a shared variance would introduce harmful features~\cite{zang2021fasa}. 
A recent work~\cite{li2021metasaug} uses a complex meta-learning strategy to learn variance of tail classes, which incurs excessive computation cost~\cite{MAML}. 

We use a simple and effective Class Adaptive Translation Coefficient mechanism. Eq.~\ref{eq8} shows $\beta_{t}$ controls the variation range of generated feature $\overline{\boldsymbol{f}}$. Since the category with fewer training instances  tends to have a smaller observed variance~\cite{liu2020deep}, $\beta_t$ is set to inversely proportional to the number of training samples $N_t$. It is formulated as:
\begin{equation}
\beta_t = \beta^0 \left(N_\text{max}-N_t\right)/\left(N_\text{max}-N_\text{min}\right),
\label{eq-b}
\end{equation}where 
$\beta^0$ is a hyperparameter, and $N_\text{max}$ and $N_{\text{min}}$ are the numbers of training samples for the most frequent and least frequent category. In this way, class adaptive $\beta_t$ can mitigate negative impact of smaller observed variance on rarer classes. 
\\

\noindent
\textbf{Logit Compensation Operation.} \quad
In Eq.~\ref{eq8}, a naive method is to explicitly translate original feature $\boldsymbol{f}$ for $M$ times, forming  an augmented feature set $\left\{\left(\overline{\boldsymbol{f}^{1}}, t\right), \left(\overline{\boldsymbol{f}^{2}}, t\right), \dots, \left(\overline{\boldsymbol{f}^{M}}, t\right)\right\}$. Then, the network can be trained with a classification loss on the augmented feature set. We consider the standard cross-entropy loss:
\begin{small}
\begin{equation}
\begin{split}
&\mathcal L_{M}(\boldsymbol{f}, \boldsymbol{W})  = \frac{1}{M}\sum_{m=1}^M-\log\left(\frac{e^{\boldsymbol{w}_{t}^T\overline{\boldsymbol{f}^{m}}}}{\sum_{k=1}^{K} e^{\boldsymbol{w}_{k}^T\overline{\boldsymbol{f}^{m}}}}\right),
\label{eq9}
\end{split}
\end{equation}
\end{small}where
$\boldsymbol{W} = [\boldsymbol{w}_1, \dots, \boldsymbol{w}_K] \in \mathbb{R}^{D\times K}$ is the weight matrix  of the linear classification layer\footnote{We set the bias vector of the classifier to $\boldsymbol{0}$.}, where $K$ is the total class number. However, the naive implementation is computationally inefficient when $M$ is large. Following ISDA~\cite{wang2019implicit}, we consider the case when $M$ grows to infinity, and derive an easy-to-compute logit compensation form for highly efficient implementation.

The augmented features can be divided into multiple groups $\{G_j\}_{j=1}^{|\mathbb{S}_t| + 1}$ based on drift directions, where the features in $G_j$ follow the Gaussian Distribution $\mathcal N \left(\boldsymbol{f} + \boldsymbol{\delta}_{tj}, \beta_{t}\boldsymbol{\sigma}_{t}^2 \right)$.
In the case of $M\to\infty$, as the probability of sampling from $\mathcal N \left(\boldsymbol{\delta}_{tj}, \beta_{t}\boldsymbol{\sigma}_{t}^2 \right)$ is $s_{tj}$, the group size of $G_j$ would be $s_{tj}M$.  We can derive the expected classification loss over the augmented feature sets as:
\begin{small}
\begin{equation}
\begin{split}
& \lim \limits_{M \to \infty} \mathcal L_{M}(\boldsymbol{f}, \boldsymbol{W})  \\
& = \lim \limits_{M \to \infty}  \frac{1}{M} \sum_{j\in \mathbb{S}_t \cup \{t\}} \sum_{\overline{\boldsymbol{f}^m} \in G_{j}} -\log \frac{e^{\boldsymbol{w}_{t}^T\overline{\boldsymbol{f}^m}  }}{\sum_{k=1}^K e^{\boldsymbol{w}_{k}^T\overline{\boldsymbol{f}^m}}} \\
& = \frac{s_{tj}M}{M} \sum_{j\in \mathbb{S}_t \cup \{t\}} \lim \limits_{M \to \infty}  \frac{1}{s_{tj}M} \sum_{\overline{\boldsymbol{f}^m} \in G_{j}} -\log\frac{e^{\boldsymbol{w}_{t}^T\overline{\boldsymbol{f}^m}  }}{\sum_{k=1}^K e^{\boldsymbol{w}_{k}^T\overline{\boldsymbol{f}^m} }} \\
& = s_{tj} \sum_{j\in \mathbb{S}_t \cup \{t\}} \mathbb{E}_{\overline{\boldsymbol{f}} \in \mathcal N \left(\boldsymbol{f} + \boldsymbol{\delta}_{tj}, \beta_{t}\boldsymbol{\sigma}_{t}^2 \right) } -\log \frac{e^{\boldsymbol{w}_{t}^T\overline{\boldsymbol{f}}   }}{ \sum_{k=1}^K e^{\boldsymbol{w}_{k}^T\overline{\boldsymbol{f}} }} \\
& = s_{tj} \sum_{j\in \mathbb{S}_t \cup \{t\}} \mathbb{E}_{\overline{\boldsymbol{f}} \in \mathcal N \left(\boldsymbol{f} + \boldsymbol{\delta}_{tj}, \beta_{t}\boldsymbol{\sigma}_{t}^2 \right) } \log  \sum_{k=1}^K e^{\left(\boldsymbol{w}_k - \boldsymbol{w}_t\right)^T \overline{\boldsymbol{f}} }. 
\end{split}
\label{eq10}
\end{equation}
\end{small}It is infeasible to tackle Eq.~\ref{eq10} directly. Following~\cite{wang2019implicit}, we use the Jensen's inequality  $\mathbb{E}[\log X] \leq \log \mathbb{E}[X]$, and the upper bound of the excepted loss can be derived as:
\begin{small}
\begin{equation}
\begin{split}
\lim \limits_{M \to \infty} \mathcal L_{M} \leq 
s_{tj} \sum_{j\in \mathbb{S}_t \cup \{t\}} \log \sum_{k=1}^K \mathbb{E}_{\overline{\boldsymbol{f}} \in \mathcal N \left(\boldsymbol{f} + \boldsymbol{\delta}_{tj}, \beta_{t}\boldsymbol{\sigma}_{t}^2 \right) }   e^{\left(\boldsymbol{w}_k - \boldsymbol{w}_t\right)^T \overline{\boldsymbol{f}}}.
\end{split}
\label{eq11}
\end{equation}
\end{small}Then
we leverage the fact that $\left(\boldsymbol{w}_k - \boldsymbol{w}_t\right)^T \overline{\boldsymbol{f}}  \sim \mathcal N \left(\left(\boldsymbol{w}_k - \boldsymbol{w}_t\right)^T \left(\boldsymbol{f} + \boldsymbol{\delta}_{tj}\right),  \beta_{t} ((\boldsymbol{w}_k - \boldsymbol{w}_t)^2)^T \boldsymbol{\sigma}_{t}^2 \right)$ and moment-generating function $\mathbb{E}[e^{X}] = e^{\mu+\frac{1}{2}\sigma}$ where $X\sim\mathcal N(\mu, \sigma)$  to obtain:
\begin{small}
\begin{equation}
\begin{split}
& \lim \limits_{M \to \infty} \mathcal L_{M} \\
&\leq s_{tj} \sum_{j\in \mathbb{S}_t \cup \{t\}} \log  \sum_{k=1}^K e^{\left(\boldsymbol{w}_k - \boldsymbol{w}_t\right)^T \left(\boldsymbol{f} +  \boldsymbol{\delta}_{tj}\right)  + \frac{\beta_{t}}{2}((\boldsymbol{w}_k - \boldsymbol{w}_t)^2)^T \boldsymbol{\sigma}_{t}^2 } \\
&= s_{tj} \sum_{j\in \mathbb{S}_t \cup \{t\}} -\log   \frac{e^{\boldsymbol{w}_{t}^T\left(\boldsymbol{f} + \boldsymbol{\delta}_{tj} \right)}}{\sum_{k=1}^K e^{\boldsymbol{w}_{k}^T\left(\boldsymbol{f} + \boldsymbol{\delta}_{tj} \right)   + \frac{\beta_t}{2}((\boldsymbol{w}_k - \boldsymbol{w}_t)^2)^T \boldsymbol{\sigma}_{t}^2}}   \\
\end{split}
\label{eq12}
\end{equation}
\end{small}Let 
$ \boldsymbol{z}_j\left(\boldsymbol{W}\right) = \boldsymbol{W} ^ T \left(\boldsymbol{f} + \boldsymbol{\delta}_{tj} \right) $ be the $K$-dimensional output logit of the linear classifier for compensated feature by FCM $\boldsymbol{f} + \boldsymbol{\delta}_{tj}$ (Eq.~\ref{eq6}). As shown in Eq.~\ref{eq12}, with LCM, the $k^{th}$ dimension of the logit  is  adjusted by $\frac{\beta_t}{2}((\boldsymbol{w}_k - \boldsymbol{w}_t)^2)^T \boldsymbol{\sigma}_{t}^2$.  Therefore, \textbf{LCM is formulated as}:
\begin{small}
\begin{equation}
\begin{split}
&\overline{\boldsymbol{z}_j}\left(\boldsymbol{W}\right) = \boldsymbol{z}_j\left(\boldsymbol{W}\right)  + \frac{\beta_t}{2}((\boldsymbol{W} - \boldsymbol{w}_t)^2)^T \boldsymbol{\sigma}_{t}^2, \quad j \in { \mathbb{S}_t \cup \{t\}}.
 \label{eq13}
\end{split}
\end{equation}
\end{small}Here,
the compensated logits $\overline{\boldsymbol{z}_j}\left(\boldsymbol{W}\right)$ can be used to compute the  classification loss for end-to-end training. The r.h.s of Eq.~\ref{eq12} can be then denoted as:
\begin{small}
\begin{equation}
\begin{split}
& \lim \limits_{M \to \infty} \mathcal L_{M}  \\
& \leq s_{tj} \sum_{j\in \mathbb{S}_t \cup \{t\}} -\log \frac{e^{\left[\overline{\boldsymbol{z}_j}(\boldsymbol{W})\right]_t}}{\sum_{k=1}^K e^{\left[\overline{\boldsymbol{z}_j}(\boldsymbol{W})\right]_k }} \\
& = \mathcal L\left(\overline{\boldsymbol{z}_j}(\boldsymbol{W}) | j \in  \mathbb{S}_t \cup \{t\} \right).
\label{eq14}
\end{split}
\end{equation}
\end{small}

\subsection{Residual Balanced Multi-Proxies  Classifier}
\textbf{Multi-Proxies Classifier.} \quad
As above analysis, the test features of one tail class tend to drift towards \textbf{multiple} similar head classes. So the distribution of test features of one tail class could present a multi-mode \cite{allen2019infinite}. In the naive linear classifier, the weight vector for each class acts as a single proxy~\cite{movshovitz2017no,douillard2020podnet}, which is difficult to optimize on a multi-mode setting. To this end, we consider a \textbf{multi-proxies} classifier to better capture the complex feature distribution. 

In particular, the multi-proxies classifier allows for $L$ proxies on each tail class during training. As pointed by~\cite{movshovitz2017no}, the proxies can be interpreted as the weight vectors in the classifier. Thus we design one weight vector for each head class, and $L$ weight vectors $\{\boldsymbol{w}_{t,l}\}_{l=1}^L$ for each tail class $t$. Then, the class-wise logit $\boldsymbol{z}\in\mathbb{R}^{K}$ for each input feature $\overline{\boldsymbol{f}}$ can be computed as,
\begin{equation}
\left[\boldsymbol{z}\right]_k  = \left \{
\begin{aligned}
&   \boldsymbol{w}_k^T\overline{\boldsymbol{f}}, \ \text{if} \ k \in C_h, \\
&  \sum_{l=1}^L \pi_{k,l}\left(\boldsymbol{w}_{k,l}^T\overline{\boldsymbol{f}}\right), \ \pi_{k,l} = \frac{e^{\boldsymbol{w}_{k,l}^T\overline{\boldsymbol{f}}}}{\sum_{i=1}^L e^{\boldsymbol{w}_{k,i}^T\overline{\boldsymbol{f}}}}, \ \text{if} \ k \in C_t,
\end{aligned}
\right.
\label{eq15}
\end{equation}where
$\pi_{k,l}$ denotes the similarity of input feature to the $l^{th}$ proxy of class $k$. Notably, the multi-proxies classifier can be regarded as a \textbf{sample-adaptive linear classifier}, where  the equivalent weight matrix is $\widehat{\boldsymbol{W}} = \left[\widehat{\boldsymbol{w}}_1, \widehat{\boldsymbol{w}}_2, \dots, \widehat{\boldsymbol{w}}_K\right]$,
\begin{equation}
\widehat{\boldsymbol{w}}_k = \left \{
\begin{aligned}
&\boldsymbol{w}_k \quad \text{if} \quad k \in C_h, \\
& \sum_{1=1}^L \pi_{k,l}\boldsymbol{w}_{k,l} \quad \text{if} \quad k \in C_l.
\label{eq16}
\end{aligned}
\right.
\end{equation}
Notably, in the case of multi-proxies classifier, LCM simply needs to replace $\boldsymbol{W}$ with $\widehat{\boldsymbol{W}}$ in Eq.~\ref{eq13}.
\\

\noindent
\textbf{Residual Balanced Multi-Proxies Classifier.} \quad
In order to  prevent the classifier predictions from highly biasing to head classes, existing methods~\cite{Decoupling,BBN} use CBS to optimize the classifier. However, CBS leads to a side-effect of making head classes under-optimized, increasing the risk of classifier under-fitting the head data. As shown in Fig.~\ref{fig2} (b), CBS classifier performs poorly on both training and test set of head classes, which proves that CBS essentially suffers from under-fitting on head classes. To this end, we design a RBMC that uses a residual mechanism to learn a class-balanced classifier.

The structure of RBMC is illustrated in Fig.~\ref{fig3}. RBMC consists of a Uniform Classifier and a Residual Classifier. Both classifiers adopt multi-proxies classifiers, which are trained under US and CBS respectively. In particular, given the feature vector $\overline{\boldsymbol{f}^u}\in\mathbb{R}^{D}$ after FCM on Uniform Branch with US, the \textbf{uniform classifier} can be formulated as:
\begin{equation} 
\boldsymbol{z}^u=\widehat{\boldsymbol{W}^u}^T\overline{\boldsymbol{f}^u},
\label{eq17}
\end{equation}where 
$\widehat{\boldsymbol{W}^u}$ denotes the weight matrix of the multi-proxies uniform classifier (Eq.~\ref{eq16}). However, since US ignores the tail data~\cite{Decoupling}, the prediction of uniform classifier is imbalanced and highly skewed to head classes. 

Therefore,  another \textbf{residual classifier} is used to produce a class balanced prediction. It learns the mapping from feature space to the residual between the imbalanced logits obtained by \textit{uniform classifier} and desired balanced logits. Given the feature vector $\overline{\boldsymbol{f}^b}\in\mathbb{R}^{D}$ after FCM on Class Balanced Branch with CBS, \textbf{RBMC is formulated as}:
\begin{equation}
\boldsymbol{z}^b=\widehat{\boldsymbol{W}^u}^T\overline{\boldsymbol{f}^b} + \widehat{\boldsymbol{W}^r}^T\overline{\boldsymbol{f}^b} = \left(\widehat{\boldsymbol{W}^u} + \widehat{\boldsymbol{W}^r}\right)^T\overline{\boldsymbol{f}^b},
\label{eq18}
\end{equation}where
$\widehat{\boldsymbol{W}^r}$ is the weight matrix of the multi-proxies residual classifier. In such design, RBMC is jointly learned under both CBS (through \textit{residual classifier}) and US (through \textit{uniform classifier}) schemes. Therefore, the head information compromised by CBS can be recovered through uniform classifier. In this sense, RBMC can utilize the information from whole training  dataset to alleviate under-fitting, while avoiding highly skewing to head classes.

\begin{algorithm}[t]
\small
\caption{Learning algorithm of DCRNets.}
\begin{algorithmic}[1]
\Require {$X$ is Training Dataset; $\mathcal F_{cnn}(\cdot;\Theta)$ is CNN feature extractor with parameter $\Theta$; $\widehat{\boldsymbol{W}^u}/\widehat{\boldsymbol{W}^r}$ is weight matrix of uniform/residual multi-proxies  classifier; $T$ is the total training epochs.}

\For{$t=1$ to $T$}
\State $\left(\mathbf{x}^u,y^u\right) \leftarrow $ $\rm{UniformSampler}\left( X \right)$
\State $\left(\mathbf{x}^b,y^b\right) \leftarrow $  $\rm{ClassBalancedSampler}\left( X \right)$
\State $\boldsymbol{f}^u \leftarrow \mathcal F_{cnn}\left(\mathbf{x}^u; \Theta \right)$
\State $\boldsymbol{f}^b \leftarrow \mathcal F_{cnn}\left(\mathbf{x}^b; \Theta \right)$
\State Send $\boldsymbol{f}^u$ to FCM to obtain $\left\{\overline{\boldsymbol{f}_j^u}\right\}_{j\in \mathbb{S}_{y^u} \cup \{y^u\}}$ with Eq.~\ref{eq6}
\State Send $\boldsymbol{f}^b$ to FCM to obtain $\left\{\overline{\boldsymbol{f}_j^b}\right\}_{j\in \mathbb{S}_{y^b} \cup \{y^b\}}$ with Eq.~\ref{eq6}
\State $\boldsymbol{z}^u_j  = \widehat{\boldsymbol{W}^u}^T\overline{\boldsymbol{f}^u_j}$ \quad (Eq.~\ref{eq17})
\State $\boldsymbol{z}^b_j  = (\widehat{\boldsymbol{W}^{u}} + \widehat{\boldsymbol{W}^{r}})^T\overline{\boldsymbol{f}^b_j}$ \quad (Eq.~\ref{eq18})
\State send $\boldsymbol{z}^u_j$ to LCM to obtain $\overline{\boldsymbol{z}^u_j}(\widehat{\boldsymbol{W}^{u}})$ with Eq.~\ref{eq13}
\State send $\boldsymbol{z}^b_j$ to LCM to obtain $\overline{\boldsymbol{z}^b_j}(\widehat{\boldsymbol{W}^{u}}+\widehat{\boldsymbol{W}^{r}})$ with Eq.~\ref{eq13}
\State $\mathcal L_1 \leftarrow \mathcal L\left(\overline{\boldsymbol{z}^u_j}(\widehat{\boldsymbol{W}^u}) | j \in \mathbb{S}_{y^u} \cup \{y^u\} \right)$ with Eq.~\ref{eq14}
\State $\mathcal L_2 \leftarrow \mathcal L\left(\overline{\boldsymbol{z}^b_j}(\widehat{\boldsymbol{W}^u} + \widehat{\boldsymbol{W}^r}) | j \in \mathbb{S}_{y^b} \cup \{y^b\} \right)$ with Eq.~\ref{eq14}
\State $\mathcal L = \phi \mathcal L_1 + (1 - \phi) \mathcal L_2$
\State Update  parameters $\{\Theta, \widehat{\boldsymbol{W}^u}, \widehat{\boldsymbol{W}^r}\}$ by minimizing $\mathcal L$
\EndFor
\end{algorithmic}
\label{algor1}
\end{algorithm}

\subsection{Overall Algorithm }

\textbf{Training phase.} \quad The training algorithm of DCRNets is shown in Algorithm~\ref{algor1}. We apply US and CBS to training set separately, and obtain sample  $\left(\mathbf{x}^u,y^u\right)$ for  uniform branch and $\left(\mathbf{x}^b,y^b\right)$ for  class balanced branch. Then, the two samples are fed into the shared feature extractor to acquire the feature vectors $\boldsymbol{f}^u\in\mathbb{R}^{D}$ and $\boldsymbol{f}^b\in\mathbb{R}^{D}$. Since the training and test features of tail categories deviate from each other, the feature vectors are sent into FCM to compensate the drift. FCM generates a set of compensated features  $\{\overline{\boldsymbol{f}^u_j}\}_{j\in \mathbb S_{y^u} \cup \{y^u\}}$ and $\{\overline{\boldsymbol{f}^b_j}\}_{j\in \mathbb S_{y^b} \cup \{y^b\}}$ (Eq.~\ref{eq6}). After that, each $\overline{\boldsymbol{f}^u_j}$ is sent into the uniform classifier of weight matrix $\widehat{\boldsymbol{W}^u}$ to predict the class-wise logit $\boldsymbol{z}^{u}_j\in\mathbb{R}^{K}$ (Eq.~\ref{eq17}). And each $\overline{\boldsymbol{f}^b_j}$ is sent into RMBC of weight matrix $\widehat{\boldsymbol{W}^u} + \widehat{\boldsymbol{W}^r}$ to predict the logit $\boldsymbol{z}^{b}_j\in\mathbb{R}^{K}$ (Eq.~\ref{eq18}). In order to prevent the model from over-fitting  to FCM errors, the output logits are adjusted by LCM and obtain compensated logits $\overline{\boldsymbol{z}^u_j}$ and $\overline{\boldsymbol{z}^b_j}$  (Eq.~\ref{eq13}). 
Finally, the classification loss on Uniform  Branch is denoted as $\mathcal L_1=\mathcal L\left(\overline{\boldsymbol{z}^u_j}(\widehat{\boldsymbol{W}^u}) | j \in \mathbb{S}_{y^u} \cup \{y^u\} \right)$ (Eq.~\ref{eq14}), and the classification loss on Class Balanced Branch is defined as  $\mathcal L_2 = \mathcal L\left(\overline{\boldsymbol{z}^b_j}(\widehat{\boldsymbol{W}^u} + \widehat{\boldsymbol{W}^r}) | j \in \mathbb{S}_{y^b} \cup \{y^b\} \right)$ (Eq.~\ref{eq14}).  The overall loss for DCRNets is computed as $\phi \mathcal L_1 + (1 - \phi) \mathcal L_2$, where $\phi$ is a trade-off hyperparameter. DCRNets can be end-to-end optimized by minimizing the overall loss function.

\noindent
\textbf{Test Phase.}  \quad In the test phase, only Class Balanced Branch is used for prediction and the compensation modules (FCM and LCM) are removed. Formally, given a test sample, we first use the feature extractor to obtain the feature vector. Then the feature is sent into RBMC to obtain a class balanced prediction (Eq.~\ref{eq18}). So DCRNets only  incur little computational burden, which is extremely efficient.

\section{Discussions}
\noindent \textbf{RBMC.} \quad The scheme of decoupling feature learning and classifier learning~\cite{Decoupling} has become a dominant direction for class-imbalanced task. Our RBMC uses uniform sampling to benefit  \textbf{classifier} learning, which effectively alleviates the under-fitting of decoupled classifier~\cite{Decoupling} to head classes. Also, the feature learning and classifier learning can obey a unified way without decoupling. The jointly learning  enhances the compatibility between features and classifier, and meets the end-to-end merit in deep learning. Considering its simplicity and effectiveness, RBMC could serve as a generic classifier in class-imbalanced learning literature.

\noindent
\textbf{FCM and LCM.} \quad Most studies on class imbalanced learning~\cite{Decoupling,BBN,hong2020disentangling} attempt to deal with \textbf{label shift} between training and test phase. Existing works generally assume that the extracted features of training and test data follow a same distribution. However, this assumption may not hold. As shown in Fig.~\ref{fig2} (a), there is a severe \textbf{feature drift} between training and test especially on tail classes. To the best of our knowledge, we are the first to point out the test features of tail classes tend to drift towards similar head classes. And our FCM and LCM can effectively alleviate the feature drift  issue through compensation operations. Also, we combine FCM and LCM with various long-tailed methods and demonstrate that the two modules can consistently boost their performances. So we suggest that FCM and LCM can be used as general modules for class imbalanced learning to enhance generalization ability.  

\noindent
 \textbf{Combination of ``FCM and LCM'' and   ``RBMC''. } \quad
In DCRNets, FCM and LCM use a set of hyper-parameters, $\{\alpha_t\}_{t=1}^K$ and $\{\beta_t\}_{t=1}^K$, to control a stronger data augmentation for tail class as compared to head class, while RBMC uses a residual path to alleviate underfitting on head class. The optimal combination of ``FCM and LCM'' and   ``RBMC'' should achieve the best trade-off performance on both head and tail classes. As shown in Eq.~\ref{eq-a} and Eq.~\ref{eq-b}, as the values of $\alpha^0$ and $\beta^0$ of FCM and LCM decrease, the difference between augmentation strength  applied to head and tail classes reduces. Thus, $\alpha^0$ and $\beta^0$ control the ability of FCM and LCM to improve tail accuracy while potentially suppressing head accuracy. So adjusting $\alpha^0$ and $\beta^0$ can modulate the ability of FCM and LCM,  enabling them to work collaboratively with RBMC to achieve the best trade-off performance on both head and tail classes.

\noindent
\textbf{\underline{Comparison to BBN} \cite{BBN}.} \quad
Our framework DCRNets share similarity to BBN~\cite{BBN} of using a two-branch architecture. However, DCRNets differ from BBN in the following ways: 

\textbf{(1) Different Network Architecture}. \textbf{From classifier view,}  the classifiers of the two branches of BBN are independent of each other. Differently, DCRBets add a \textit{residual connection} from uniform classifier to class-balanced classifier. With the residual connection, the class-balanced classifier can effectively utilize the prediction of uniform classifier, thereby better fitting the head data;  \textbf{From feature extractor view,} BBN uses \textit{separate} parameters for part of feature extractors of the two branches. Instead, DCRNets \textit{share} the backbone parameters, which largely reduces computation cost in the inference phase. 

\textbf{(2) Different Learning Mechanism.} BBN adopts \textit{cumulative learning} mechanism while our DCRNets adopt \textit{residual learning} mechanism. The cumulative learning of BBN gradually assigns a bigger weight to class-balanced branch. However, as pointed by \cite{Decoupling}, over-emphasizing on class balanced learning even only at the later training stage damages the generalization of feature representation. Differently, DCRNets use a residual connection to bridge uniform learning and class-balanced learning, and always gives a fixed and dominant weight to uniform learning branch, which ensures high-quality representation.  

\textbf{(3) DCRNets add FCM and LCM modules.} BBN suffers from feature drift between training and test data, which leads to overfitting on tail classes. On the contrary, FCM and LCM in DCRNets can effectively alleviate the feature drift, thereby achieving a more generalizable classifier. In addition, FCM and LCM can applied to various long-tailed learning methods including one-branch~\cite{Decoupling} and two-branch methods~\cite{BBN}. Therefore, different from BBN, our proposed method is not limited to a two-branch architecture.

\noindent
\textbf{\underline{Comparison to ResLT}~\cite{cui2022reslt}.} \quad
Our RBMC and ResLT~\cite{cui2022reslt} share a similar idea of using a residual mechanism for long-tailed recognition. However, they have following sufficient differences: 

(1) \textbf{The implementation and function of residual learning are fundamentally different}. ResLT~\cite{cui2022reslt} designs three branches  equipped with many+medium+tail, medium+tail and tail data respectively. It adds the residual connections \textit{from medium+tail/tail branch to main branch}, which aims to \textit{enhance performance on medium and tail data}. On the contrary, our RBMC designs two branches  equipped with uniform and class-balanced sampling respectively. RBMC adds a residual connection \textit{from uniform branch to class-balanced branch}, which aims to \textit{enhance performance on head data}.

(2) \textbf{The architecture of residual branches is different}. ResLT conducts re-balance in the aspect of parameter space  of \textit{feature extractor}. So ResLT allocates individual parameters for feature extractor and uses a shared classifier on  residual branches. In contrast, RBMC conducts re-balance in the aspect of parameter space of \textit{classifier}. So RBMC uses a shared feature extractor and allocates individual parameters for classifier on residual branch.

(3) \textbf{The learning strategy of residual branches is different}. ResLT optimizes residual branches using different sub-group data, \textit{i.e.}, medium+tail and tail data, while RBMC optimizes residual branch with a class-balanced sampling strategy. Therefore, RBMC does not need to artificially divide many, medium and tail data, which is more flexible and easier to implement

\noindent
\textbf{\underline{Comparison to ISDA} \cite{wang2021regularizing}.} \quad
Recently, \cite{wang2021regularizing} proposes a similar semantic data augmentation algorithm ISDA. We summarize the main differences between ISDA and LCM. 

(1) Problem Setting: ISDA mainly focuses on a \textit{balanced} classification task while our work considers the problem of learning form long-tailed \textit{imbalanced} dataset. 

(2) Insight: ISDA aims to \textit{augment} training samples to regularize deep networks. Differently, our LCM aims to better \textit{calibrate} the training features to the corresponding test feature space, which aims to learn a more generalizable classifier for tail classes. 

(3) Technique: In order to better address long-tailed problem, LCM has the following different designs from ISDA. \textbf{Firstly, LCM performs augmentation around the feature drift vectors instead of original features.} As shown in Equation 4 of \cite{wang2021regularizing}, \textit{i.e.}, $\widetilde{\boldsymbol{a}}_i \sim \mathcal N \left(\boldsymbol{a}_i, \lambda \Sigma_{y_i} \right)$, ISDA performs augmentation around the \textit{original features}.  But on long-tailed recognition, due to the scare samples of tail classes, there is a large drift between training and test features. So the augmented features around original training features would probably not lie in true (test) feature manifold, especially for tail classes. Differently, as shown in Eq.~\ref{eq7}, LCM performs augmentation around the \textit{feature drift vector} $\boldsymbol{\delta}_{tj}$, in which $\boldsymbol{\delta}_{tj}$ is crucial to bridge the feature drift of tail classes to  generate more realistic tail features. 

\textbf{Secondly, LCM derives derivation based on Gaussian Mixture Distribution instead of Gaussian Distribution.} As shown in Equation 7 of \cite{wang2021regularizing}, ISDA derives the derivation based on a \textit{Gaussian Distribution}. It is built on the assumption that the feature representation of each class follows a \textit{uni-modal} Gaussian distribution. However, as pointed by \cite{allen2019infinite}, the feature distribution of test data of a tail class usually presents \textit{multi-modal} \cite{allen2019infinite}.
To this end, LCM takes a step further and extends the derivation to \textit{Gaussian Mixture Distribution} (Eq.~\ref{eq10}). Built on Gaussian Mixture Distribution, LCM is more conducive to capture the complex multi-modal  distributions of tail classes. 

\textbf{Thirdly, LCM adapts Class Adaptive Translation instead of same augmentation for each class.} As shown in Equation 4 of \cite{wang2021regularizing}, \textit{i.e.}, $\widetilde{\boldsymbol{a}}_i \sim \mathcal N \left(\boldsymbol{a}_i, \lambda \Sigma_{y_i} \right)$, ISDA uses the same augmentation strength ($\lambda$) for all classes. However, on long-tailed recognition, the observed variance of tail class is usually very small~\cite{liu2020deep}. Therefore, the augmented tail features of ISDA would be in closer vicinity to the original ones, thereby limiting the tail data variety and even aggravating imbalance issue. Differently, as shown in Eq.~\ref{eq7}, LCM designs a \textit{class adaptive translation coefficient} ($\beta_t$) to adaptively increase the augmentation strength for tail classes, thereby ensuring the diversity of generated tail features.

\section{Experiments}
We mainly evaluate the proposed DCRNets on long-tailed classification task. We use five long-tailed benchmarks, \textsl{i.e.}, ImageNet-LT \cite{imageNEt}, iNaturalist18 \cite{van2018inaturalist}, CIFAR100-LT \cite{krizhevsky2009learning}, CIFAR10-LT \cite{krizhevsky2009learning} and PLACE-LT \cite{imageNEt}. We incorporate DCRNets with  multiple network architectures (ResNet and ResNeXt), and a multi-expert framework and achieve the best model performance for all settings. We also evaluate DCRNets on Class Incremental Learning task on CIFAR100 \cite{krizhevsky2009learning} to show its generality.

\subsection{Datasets and Evaluation Protocol}
\textbf{ImageNet-LT dataset.} \quad
ImageNet-LT is proposed by~\cite{imageNEt}. ImageNet-LT is a long-tailed subset of ImageNet-2012~\cite{deng2009imagenet} by sampling a subset following the Pareto distribution with power value of 6. It contains 105.8K training images and 50,000 test images from 1,000 categories. The number of samples per class ranges from 5 to 1280. 

\textbf{iNaturalist18 dataset.} \quad
iNaturalist18 \cite{van2018inaturalist} is one species classification dataset with images collected from real world, which is on a large scale and suffers from extremely imbalanced class distribution. It is composed of 437.5K training images and 2.4K test images from 8,142 categories. The number of samples per class ranges from 2 to 1,000. In addition to the extreme imbalance, iNaturalist18 also confronts the fine-grained problem.

\textbf{CIFAR100-LT and CIFAR10-LT dataset.} \quad
CIFAR10 and CIFAR100 dataset have 60,000 images, 50,000 for training and 10,000 for validation with 100 and 10 categories. Following~\cite{cao2019learning,BBN}, we use a long-tailed version of CIFAR10/CIFAR100 dataset, which follows an exponential decay in sample sizes across different classes with various imbalance factors. The imbalance factor is defined as the ratio between the numbers of training samples for the most frequent class and the least frequent class. We use the imbalance factor of $100$ in our experiments.

\textbf{Places-LT dataset.} \quad
Places-LT is proposed by~\cite{imageNEt}. Places-LT is a long-tailed version of the large-scale scene classification dataset Places~\cite{zhou2014learning}. It consists of 184.5K images from 365 categories. The number of samples per class ranges from 5 to 4980.

\textbf{Evaluation Protocol.} \quad
After training on long-tailed dataset, we evaluate the models on a balanced test set, and report the top-1 accuracy over all classes, denoted as ``All''. Following~\cite{imageNEt}, we further report accuracy on three  splits of classes: \textit{Many-shot} (more than 100 samples), \textit{Medium-shot} (with 20 to 100 samples) and \textit{Few-shot} (less than 20 samples).

\subsection{Implementation Details} 
\textbf{Implementation Details on ImageNet-LT.}  \quad 
For ImageNet-LT, we use standard ResNet-10, ResNet-50~\cite{residual} and ResNeXt-50~\cite{xie2017aggregated} as backbone. We follow the same training strategy in~\cite{Decoupling}. In details, we use SGD optimizer with momentum 0.9 and train for 90 and 200 epochs respectively. We adopt cosine learning rate schedule gradually decaying from 0.075 to 0, and batch size 192 and 64 for uniform branch and class balanced branch respectively. We define the classes with more than 100 training samples as head classes, and the remaining classes as tail classes. In FCM, the number of selected head classes $m$ is 2 (Eq.~\ref{eq1}). The parameter $\alpha^0$ is set to  $0.5$/$0.5$/$1.0$ for ResNet-10/ResNet-50/ResNeXt-50. In LCM, the parameter $\beta^0$ is set to $1.5$/$6.0$/$9.0$ for ResNet-10/ResNet-50/ResNeXt-50. In RBMC, the number of proxies on each tail class ($L$) is set to 2 (Eq.~\ref{eq15}). The weighting factor of loss function $\phi$ is set to 0.8.

\textbf{Implementation Details on iNaturalist18.} \quad
Following~\cite{hong2020disentangling}, we utilize ResNet-50 as backbone and train it for 90 and 200 epochs respectively. We set $\alpha^0$ to $0.5$, $\beta^0$ to $3.0$.  Other settings are the same as those in ImageNet-LT.

\textbf{Implementation Details on CIFAR100-LT.}\quad
For CIFAR-100 dataset, the implementation details mainly follow~\cite{Balanced}. We use ResNet32 as backbone. The networks are trained for 500 epochs. The initial learning rate is set to 0.05, and the batch size of uniform branch and class balanced branch is set to 384 and 128 respectively. For the hyperparameter, $\alpha^0$ and $\beta^0$ are set to $0.5$ and $1.0$ respectively. Other setting are the same as those in ImageNet-LT.

\textbf{Implementation Details on Places-LT.}\quad 
For Places-LT, following previous setting~\cite{BBN}, we choose ResNet-152 pre-trained on ImageNet dataset as backbone network, and train the model for 30 epochs. We set $\alpha^0$ to 0.5, $\beta^0$ to 1.0.  Other settings are the same as those in ImageNet-LT.

\subsection{Comparisons with State-of-the-art Methods}
In this section, we compare our DCRNets with other state-of-the-arts  on ImageNet-LT, iNaturalist18, CIFAR100-LT and Places-LT benchmarks. Numerical results can be found in Tab.~\ref{Tab1}-\ref{Tab5}.

The compared methods cover various categories of ideas for imbalanced classification:
\begin{enumerate}[]
\item \textbf{Re-weighting}  (\textbf{RW}). Re-weighting the loss function, such as Focal Loss~\cite{lin2017focal}, BALMS~\cite{Balanced} and LDAM~\cite{cao2019learning}.
\item \textbf{Decoupled  scheme} (\textbf{DE}). Decoupling the feature learning and classification, such as Two-stage~\cite{Decoupling,zhang2021distribution,tang2020long,hong2020disentangling}, CBS+RRS~\cite{zhang2019balance} and BBN~\cite{BBN}.
\item \textbf{Data Generation} (\textbf{DG}). Generating new samples for tail categories, such as MetaSAug~\cite{li2021metasaug} and RSG~\cite{wang2021rsg}.
\item \textbf{Transfer Learning} (\textbf{TL}). Transferring knowledge learned from head classes to tail classes, such as OLTR~\cite{imageNEt} and M2m~\cite{kim2020m2m}.
\item \textbf{Representation Learning} (\textbf{RL}). Using self-supervised for better representation: SSP~\cite{SSL}, Hybrid-PSC~\cite{wang2021contrastive} and  PaCo~\cite{cui2021parametric}.
\item \textbf{Multi-Expert Network} (\textbf{ME}). Using an ensemble network to extract features  or extra distillation training procedures: RIDE~\cite{wang2020long}, ACE~\cite{cai2021ace},  ResLT~\cite{cui2022reslt} and NCL~\cite{li2022nested}.
\end{enumerate}

\begin{table}[t]
\centering
\small
\caption{Long-tailed classification performance (overall top-1 accuracy $\%$) compared to related methods on \textbf{ImageNet-LT}. $\dag$ indicates our reproduced results with the released code. R-10, R-50 and X-50 means ResNet-10, ResNet-50 and ResNeXt-50.}
\begin{center}
\begin{tabular}{c | c | l |  l | l | l   }
\hline
 Epoch & \multicolumn{2}{c|}{Method}  & R-10  & R-50 &RX-50 \\   
\hline
\multirow{13}*{90} & \multirow{2}*{RW} 
& Focal~\cite{lin2017focal} &30.5 & - &43.7 \\
& & BALMS~\cite{Balanced} & 41.8 & 48.8$\dag$ &51.4 \\
\cline{2-6}
& \multirow{7}*{DE} 
&$\tau$-norm~\cite{Decoupling} & 40.6 &46.7 & 49.4\\
& & cRT~\cite{Decoupling} & 41.8 & 47.3  & 49.5   \\
& & CBS+RRS~\cite{zhang2019balance}& 41.9  &47.3  &-\\
& & LWS~\cite{Decoupling} & 41.4 &47.7 &49.9   \\
& & LADE~\cite{hong2020disentangling} &41.6$^{\dag}$   &50.8$^{\dag}$ &51.9   \\
& &TDE~\cite{tang2020long} &- & 51.1&51.8\\
& & DisAlign~\cite{zhang2021distribution} & - & 51.3 &52.6 \\
\cline{2-6}
& \multirow{2}*{DG} 
& MetaSAug~\cite{li2021metasaug} &-&  47.4&-\\
& & RSG~\cite{wang2021rsg} &- &- &51.8\\
\cline{2-6}
& \multirow{2}*{TL} 
& M2m~\cite{kim2020m2m} &- &43.7 &-\\
& & OLTR~\cite{imageNEt} &37.3 & - &46.3 \\
\cline{2-6}
& 
& \textbf{DCRNets} &\textbf{43.8} &\textbf{53.2} &\textbf{54.8}\\
\hline
\multirow{5}*{ $>$ 180} & \multirow{3}*{DE}
& cRT~\cite{Decoupling} & 42.7$^{\dag}$ & 50.8$^{\dag}$  &  51.2$^{\dag}$  \\
& & LADE~\cite{hong2020disentangling} &43.1$^{\dag}$   &52.3$^{\dag}$ &53.0  \\
& & LWS~\cite{Decoupling} & 43.5$^{\dag}$ & 51.7$^{\dag}$ & 51.9$^{\dag}$  \\
\cline{2-6}
& \multirow{1}*{RL}
&SSP~\cite{SSL} &43.2 &51.3 &-\\
\cline{2-6}
&\multirow{1}*{ME}
& ResLT~\cite{cui2022reslt} &  43.8 &- & 52.9 \\
\cline{2-6}
& & \textbf{DCRNets} &\textbf{45.0} &\textbf{54.1} &\textbf{55.0}\\
\hline
\hline
\multirow{5}*{ 100} & \multirow{5}*{ME}
&RIDE~\cite{wang2020long}(2 experts) & 45.3 & 54.4 & 55.9\\
& &\textbf{DCRNets}(2 experts) & \textbf{46.6} & \textbf{55.7} &\textbf{56.8}\\
\cline{3-6}
& & ACE~\cite{cai2021ace}(3 experts) & 44.0 & 54.7 & 56.8\\
& &RIDE~\cite{wang2020long}(3 experts) & 45.9 & 54.9 & 56.4\\
& &\textbf{DCRNets}(3 experts) & \textbf{47.6} &\textbf{56.9} &\textbf{57.2}\\
\hline
\end{tabular}
\end{center}
\label{Tab1}
\end{table}

\begin{table}[t]
\centering
\small
\captionsetup{font={small}}
\caption{Comparison to related methods on \textbf{iNaturalist18}.}
\begin{center}
\begin{tabular}{c | l  | l l }
\hline
\multicolumn{2}{c|}{Method}  & 90E & $>$180E \\ 
\hline
\multirow{3}*{RW} 
&Focal~\cite{lin2017focal}  & 61.1 &- \\
&LDAM~\cite{cao2019learning}  1 & 64.6 &-\\
&BALMS~\cite{Balanced}  & - &69.8\\
\hline
\multirow{7}*{DE}
&cRT~\cite{Decoupling}  &65.2 &67.6\\
&$\tau$-norm~\cite{Decoupling}  & 65.6 &69.3\\
&LWS~\cite{Decoupling}  & 65.9 &69.5\\
&BBN~\cite{BBN}  & 66.3 & 69.3\\
&LADE~\cite{hong2020disentangling}  &- &70.0\\
&DisAlign~\cite{zhang2021distribution}  &67.8 & 70.6\\
&LDAM+DRW~\cite{cao2019learning}  &68.0 &-\\
\hline
\multirow{2}*{DG}
&MetaSAug~\cite{li2021metasaug}  & 66.3 &-\\
&RSG~\cite{wang2021rsg}  &67.9 &70.3 \\
\hline
\multirow{2}*{RL}
&SSP~\cite{SSL}   &- &68.1\\
&Hybrid-PSC~\cite{wang2021contrastive} & 68.1 & 70.4\\
\hline
\multirow{1}*{ME}
& ResLT\cite{cui2022reslt} & - & 70.2\\
\hline
& \textbf{DCRNets}  &\textbf{70.3} &\textbf{72.3} \\
\hline
\hline
\multirow{6}*{ME}
&RIDE~\cite{wang2020long} (2 experts)  & 71.4 &- \\
& \textbf{DCRNets} (2 experts)&\textbf{72.0} &-\\ 
\cline{2-4}
&RIDE~\cite{wang2020long} (3 experts) & 72.2 &- \\
&ACE~\cite{cai2021ace} (3 experts) & 72.9 &-\\
& \textbf{DCRNets} (3 experts)&\textbf{73.0} &-\\
\cline{2-4}
\hline
\end{tabular}
\end{center}
\label{Tab2}
\end{table}

 \begin{table}[t]
\centering
\small
\captionsetup{font={small}}
\caption{Comparison to related methods on \textbf{CIFAR100-LT}  and \textbf{CIFAR10-LT} with an imbalance of 100.}
\begin{center}
\begin{tabular}{c | l |  c | c }
\hline
\multicolumn{2}{c|}{Method} & CIF100-LT & CIF10-LT \\
\hline
\multirow{3}*{RW} 
&Focal~\cite{lin2017focal} & 38.4 &70.4 \\
&LDAM~\cite{cao2019learning} &42.1 &77.0\\
&BALMS~\cite{Balanced} &50.8 &84.9\\
\hline
\multirow{4}*{DW}
&BBN~\cite{BBN} & 42.6 & 79.8\\
&Logit Adj.~\cite{menon2020long} & 43.9 & 77.7\\
&cRT~\cite{Decoupling} &50.0 &82.0 \\
&LWS~\cite{Decoupling} &50.5 &83.7 \\
\hline
\multirow{2}*{DG}
&RSG~\cite{wang2021rsg} &44.6 &79.6\\
&MetaSAug~\cite{li2021metasaug} &48.0 &80.7\\
\hline
\multirow{2}*{TL}
&OLTR~\cite{imageNEt} & 41.2 &-\\
&M2m~\cite{kim2020m2m} &43.5 & 79.1\\
\hline
\multirow{2}*{RL}
&Hybrid-PSC~\cite{wang2021contrastive} & 44.9 &78.8\\
&Hybrid-PSC~\cite{wang2021contrastive} & 46.7 & 81.4 \\
\hline
\multirow{1}*{ME}
& ResLT~\cite{cui2022reslt}&  45.3 & 80.4\\
\hline
\multirow{4}*{}
& \textbf{DCRNets} &\textbf{51.4} &\textbf{85.0}  \\
\hline
\hline
\multirow{6}*{ME}
&RIDE~\cite{wang2020long} (2 experts) & 47.0 &- \\
& \textbf{DCRNets} (2 experts)&\textbf{55.0} &\textbf{85.8}\\
\cline{2-4}
&RIDE~\cite{wang2020long} (3 experts) & 48.0 &- \\
&ACE~\cite{cai2021ace} (3 experts)& 49.4 &81.2\\
& \textbf{DCRNets} (3 experts) &\textbf{56.0} &\textbf{87.0}\\
\hline
\end{tabular}
\end{center}
\label{Tab3}
\end{table}

\textbf{Main Results on ImageNet-LT.} \quad Tab.~\ref{Tab1} reports the main results on ImageNet-LT. Our method achieves the best performance under the same setting. It is noted that: \textbf{(1)}  Decoupled methods 
\cite{Decoupling,hong2020disentangling,zhang2021distribution,tang2020long} (DE) are prone to under-fitting the head and over-fitting the tail classes. It is noteworthy that DCRNets have outperformed all the decoupled methods by about $2\%$ top-1  accuracy, which validates its superiority. \textbf{(2)} Data-generation methods \cite{li2021metasaug,wang2021rsg} (DG)  and Transfer-learning methods  \cite{kim2020m2m,imageNEt} (TL) use complex generation/transfer modules and explicitly generate features. By contrast, DCRNets implicitly generate features by two shift operations. Our method puts much less overhead with a much better performance: about $3\%$ top-1 accuracy improvement. \textbf{(3)} Representation-learning method \cite{SSL} (RL) employs a self-supervised learning to obtain a good feature initialization. DCRNets still surpass SSP \cite{SSL} by $1.8\%$ top-1 accuracy with ResNet10 as backbone and $2.8\%$ top-1 accuracy with ResNet50 as backbone. Notably, as a model initialization technique, SSP is orthogonal to our method and can be easily combined to further improve performance.  \textbf{(4)} ResLT~\cite{cui2022reslt} and our method both use a residual mechanism for long-tailed recognition. Our method surpasses ResLT by $1.2\%$ and $2.1\%$ top-1 accuracy with ResNet10 and ResNeXt50 as backbone respectively, validating its superiority.

 Recently, Multi-Expert networks \cite{wang2020long,cai2021ace} (ME) achieve state-of-the-art performance on long-tailed recognition. For fair comparison, we further adopt the multi-expert network of RIDE \cite{wang2020long} as backbone. Notably, the self-distillation and EA module in RIDE are not used for simplification. As shown in Tab.~\ref{Tab1}, under the same multi-expert network, our method outperforms RIDE \cite{wang2020long} and ACE \cite{cai2021ace} by $1\%-2\%$ top-1 accuracy, which shows the proposed modules can well incorporate with multi-expert networks. 

 \textbf{Main Results on iNaturalist18.} \quad Tab.~\ref{Tab2} reports the main results on a real world benchmark, iNaturalist18. Our method also achieves the best accuracy. Compared to non-Multi-Expert methods, DCRNets outperform the best Hybrid-PSC \cite{wang2021contrastive} by about $2\%$ top-1 accuracy. Under the multi-expert network, DCRNets still outperform RIDE \cite{wang2020long} and ACE \cite{cai2021ace}. The improvements prove that our proposed method consistently improves performance on both artificial and real-world  large-scale datasets.

\begin{table}[t]
\centering
\small
\caption{Comparison to related methods on \textbf{Places-LT}.}
\begin{center}
\begin{tabular}{c | l | c  }
\hline
\multicolumn{2}{c|}{Method} & ResNet-152 \\
\hline
\multirow{3}*{RW} 
&Focal~\cite{lin2017focal} & 34.6 \\
&BALMS~\cite{Balanced} &38.6\\
&LADE~\cite{cao2019learning} & 38.8 \\
\hline
\multirow{4}*{DE}
&cRT~\cite{Decoupling} & 36.7\\
&LWS~\cite{Decoupling} & 37.6\\
& $\tau$-norm~\cite{Decoupling} &37.9 \\
&LADE~\cite{hong2020disentangling} &38.8\\
\hline
\multirow{1}*{DG}
&RSG~\cite{wang2021rsg} & 39.3\\
\hline
\multirow{1}*{TL}
&OLTR~\cite{imageNEt} & 35.9\\
\hline
\multirow{1}*{ME}
& ResLT~\cite{cui2022reslt} &  39.8\\
\hline
\multirow{2}*{}
& \textbf{DCRNets} &\textbf{40.0} \\
\hline
\end{tabular}
\end{center}
\label{Tab4}
\end{table}

\begin{table*}[t]
\centering
\small
\caption{Comparison with PaCo~\cite{cui2021parametric} and NCL~\cite{li2022nested}. We use ResNet-50 as backbone network for ImageNet-LT, and use the same training strategies as \cite{cui2021parametric,li2022nested}: training all the models  with RandAugment~\cite{cubuk2020randaugment} for 400 epochs except models on Places-LT, which is 30 epochs. ``RA'' denotes RandAugment, ``DIS'' denotes Distillation and ``SS'' denotes self-supervised learning.}
\vspace{-3.5mm}
\begin{center}
\begin{tabular}{l | l | c c c | c | c | c c | c  }
\hline
& Method   & RA & DIS  & SS& Img-LT &iNat18 & CIF100-LT & CIF10-LT & Places-LT    \\
\hline
\multirow{3}*{Single Model}  
& PaCo~\cite{cui2021parametric}  & $\checkmark$ & $\times$&$\checkmark$ & 57.0 &73.2 &\textbf{52.0} &- & 41.2  \\
& \textbf{DCRNets}   & $\checkmark$ &$\times$ & $\times$ & \textbf{57.6} &\textbf{74.0} &\textbf{52.0} &85.1 &\textbf{41.5} \\
\cline{2-10}
& \textbf{DCRNets+PaCo}   & $\checkmark$ &$\times$ & $\checkmark$ & \textbf{58.0} &\textbf{74.2} &\textbf{52.2} &\textbf{85.5} &\textbf{41.7} \\
\hline
\multirow{4}*{Multiple Models} 
& NCL~\cite{li2022nested} (single)  & $\checkmark$ & $\checkmark$& $\checkmark$& 57.4 &74.2 & 53.3 & 84.7 & 41.5\\
& \textbf{DCRNets} (single)   & $\checkmark$ &$\checkmark$ &$\times$ & \textbf{59.0} &\textbf{74.8} &\textbf{54.0} &\textbf{85.9} &\textbf{41.6}\\
\cline{2-10}
& NCL~\cite{li2022nested} (ensemble)   & $\checkmark$ & $\checkmark$& $\checkmark$&59.5 &74.9 &54.2 &85.5 &41.8\\
& \textbf{DCRNets} (ensemble)  & $\checkmark$  &$\checkmark$ &$\times$ & \textbf{60.3} &\textbf{75.7}  & \textbf{56.0} & \textbf{87.0} &\textbf{42.0}\\
\hline
\end{tabular}
\end{center}
\label{Tab5}
\end{table*}

\textbf{Main Results on CIFAR100-LT and CIFAR10-LT.} \quad
 Tab.~\ref{Tab3} \quad reports the main results on  CIFAR100-LT and CIFAR10-LT. Our method significantly outperforms the existing methods. By introducing the powerful compensation modules without using multi-expert, DCRNets are even competitive compared with multi-expert methods. Combing with multi-expert networks, DCRNets surpass RIDE and ACE by up to $6\%$ top-1 accuracy. The superior results on the two small-scale datasets further validate the effectiveness and robustness of our method.

\textbf{Main Results on Places-LT.} \quad
Tab.~\ref{Tab4}  reports the main results on  Places-LT dataset. DCRNets yields $40.0\%$ top-1 accuracy, with a notable performance gain at $0.2\%$ over the prior methods. Places-LT has an extremely imbalance class distribution, where the imbalance factor is up to 990. The superior results on this highly long-tailed benchmark further validate that our method can effectively address the class imbalance issue.

\subsubsection{Comparison with PaCo and NCL.}
Recently, PaCo~\cite{cui2021parametric} and NCL~\cite{li2022nested} achieve state-of-the-art performance on long-tailed recognition. PaCo~\cite{cui2021parametric} improves supervised contrastive loss  by adding a set of parametric learnable class centers to tackle imbalance issue. PaCo aims to utilize self-supervised contrastive learning to improve \textit{\textbf{feature representation}}. Differently, our method employs classification learning, which uses residual connection, feature shift and logit shift operations to learn a more generalizable \textit{\textbf{classifier}}. So PaCo is complementary to our method, which can be easily combined to learn better feature representation; NCL~\cite{li2022nested} is built on a multi-expert framework, which learns multiple experts concurrently and uses knowledge distillation to enhance each single expert. Our method can be integrated into NCL framework,  where we use DCRNets as each expert of NCL framework. To sum up, the techniques of PaCo and NCL are complementary to our method, which can be easily incorporated to further improve performance.

In implementation, PaCo and NCL both use the strong data augmentation RandAugment~\cite{cubuk2020randaugment}, which is well known to be a practical strategy to improve performance. And NCL uses multi-expert and knowledge distillation to further improve classification accuracy. So for a fair comparison, we use the same training strategies as \cite{cui2021parametric,li2022nested}, \textit{i.e.}, training all the models with RandAugment~\cite{cubuk2020randaugment} for 400 epochs except models on Places-LT, which is 30 epochs\footnote{Notably, NCL~\cite{li2022nested} did not use RandAugment~\cite{cubuk2020randaugment} on Places-LT benchmark. For a fair comparison on Places-LT, we reimplement NCL  with RandAugment on Places-LT, and report this result on Tab.~\ref{Tab5}.}. To compare with NCL, we also apply multi-expert and knowledge distillation strategies to our method. The results are listed in Tab.~\ref{Tab5}.

\textbf{Compared to PaCo~\cite{cui2021parametric}.} \ As shown in Tab.~\ref{Tab5}, our method can be well incorporated with strong augmentation. Equipped with  RandAugment~\cite{cubuk2020randaugment}, our method achieves $57.6\%$ and $74.0\%$ top-1 accuracy on ImageNet-LT and iNaturalist18, outperforming PaCo by $0.6\%$ and $0.8\%$ respectively. Furthermore, we add PaCo loss to our method. As shown in Tab.~\ref{Tab5}, integrating with PaCo can further lift the performance by $0.2\%-0.4\%$, which shows our method can well combine with self-supervised contrastive  PaCo.

\textbf{Compared to NCL~\cite{li2022nested}.} \ NCL adopts a multi-expert framework. NCL maintains three experts and uses a \textbf{distillation} training phase, which increases the training computational cost by almost three times. For a fair comparison, we also train three separate DCRNet experts and add a simple Kullback-Leibler (KL) distillation loss~\cite{zhang2018deep} among any two experts. Notably, the Hard Category Mining and Self-supervised contrastive loss in NCL are not used in our method.  Tab.~\ref{Tab5} reports the performance of a single expert and an ensemble of multiple experts following \cite{li2022nested}. As seen, our method can be well incorporated with distillation technique to further improve accuracy. \textit{\textbf{Under the same multi-expert distillation framework}}, our method outperforms NCL by $1.6\%$, $0.6\%$, $0.7\%$, $1.2\%$ (using a single expert) and $0.8\%$, $0.8\%$, $1.8\%$, $1.5\%$ (using an ensemble of experts) on imageNet-LT, iNaturalist18, CIFAR100-LT and CIFAR10-LT benchmarks respectively, and  achieves comparable performance on Places-LT benchmark, showing the superiority of our method. 

\subsection{Ablation Study}
\begin{table}[t]
\centering
\scriptsize
\caption{Ablation study on ImageNet-LT and iNaturalist18 for 90 epochs training.}
\begin{center}
\begin{tabular}{c | c c c | c  c c | c }
\hline
\multirow{2}*{Method} & \multirow{2}*{RBMC} & \multirow{2}*{FCM}& \multirow{2}*{LCM} & \multicolumn{3}{c|}{ImageNet-LT} &\multirow{2}*{iNat} \\
\cline{5-7}
& & & & R10 &R50 &RX50 \\
\hline
cRT~\cite{Decoupling} & & & &41.8 &47.3 &49.5 & 65.2 \\
\hline
 &$\checkmark$ & & &42.8 &51.4 &52.9&67.7\\
 &$\checkmark$ &$\checkmark$ & &43.4 &52.3 &54.0 &69.2\\
 \hline
 DCRNets &$\checkmark$ &$\checkmark$ &$\checkmark$ &\textbf{43.8} &\textbf{53.2} &\textbf{54.8} & \textbf{70.3}\\
\hline
\end{tabular}
\end{center}
\label{Tab6}
\end{table}

To investigate the effectiveness of each component in DCRNets, we conduct a series of ablation studies on ImageNet-LT and iNaturalist18. All models are trained for 90 epochs in this section. Tab.~\ref{Tab6} and Tab.~\ref{Tab7} summary the comparison results for different  settings.

\noindent
\textbf{Two-stage Learning \textit{v.s.} Joint Learning.} \quad
In order to verify the effectiveness of RBMC, we provide three variants of RBMC: \textbf{RBMC-wo-R}, \textbf{RBMC-wo-MC} and \textbf{RBMC-wo-R\&MC}. RMBC-wo-R uses multi-proxies classifiers without residual connection; RBMC-wo-MC uses linear classifiers with a residual connection; RBMC-wo-R\&MC uses linear classifiers without residual connection. We first investigate the necessity of joint learning of feature extractor and classifier. To verify this, we choose the two-stage decoupled cRT~\cite{Decoupling} as a baseline. As shown in Tab.~\ref{Tab7}, all variants of RBMC achieve obviously superior performance compared to cRT. We argue that the two-stage training of cRT harms the compatibility between features and classifier, resulting in poor adaptability of classifier. Therefore, it is necessary to jointly optimize the feature extractor and classifier for class imbalanced learning.
\\

\noindent
\textbf{Effectiveness of Residual Connection in RBMC.} \quad
We further assess the effectiveness of the residual connection in RBMC. As shown in Tab.~\ref{Tab7},  compared to RBMC-wo-R\&MC, employing residual mechanism (RBMC-wo-MC) brings $1.1\%$ overall gain with negligible computational overhead. Also, compared to RBMC-wo-R, RBMC achieves $1.0\%$ overall gain, which validates the capability of the residual connection.
Specifically, employing residual connection brings above $3\%$ gain on many-shot split. The reason is that RBMC-wo-R directly applies CBS to train the classifier of class balanced branch, which leads to under-fitting on head classes. On the contrary, RBMC introduces a residual path from uniform classifier. It ensures the class balanced branch can exploit the rich information of head classes encoded by uniform classifier. Such that the under-fitting issue can be alleviated and a better accuracy on many-shot split could be achieved. However, RMBC-wo-R achieves better performance on few-shot split. It is reasonable since the classifier of RMBC-wo-R puts more focus on few-shot data. But the overall $1.0\%$ accuracy gain manifests that RMBC derives a more balanced decision boundary which better trade-offs the head and tail classes.
\\

\noindent
\textbf{Effectiveness of Multi-proxies Classifier in RBMC.} \quad
We then investigate the influence of multi-proxies classifier (MC) in RBMC. As shown in Tab.~\ref{Tab7},  Compared to RBMC-wo-R\&MC, employing MC (RBMC-wo-R) brings $0.5\%$ overall gain. And compared to RBMC-wo-MC, RBMC achieves $0.4\%$ overall gain with negligible computational overhead. The consistent improvements validate the effectiveness of MC.
Specifically, MC achieves above $1.0\%$ improvement  on few-shot split, which indicates the robustness of MC on tail classes. Notably, RBMC shares a similar idea with Infinite Mixture Prototypes (IMP)~\cite{allen2019infinite} that represents each few-shot class by a set of clusters. However, IMP only considers the closest cluster of each class, while MC considers all clusters of a class. That allows MC to decrease the intra-class similarity when the training instances are insufficient, thereby alleviating the overfitting to the few training instances of tail classes.
\\
 
 \begin{table}[t]
\centering
\scriptsize
\captionsetup{font={small}}
\caption{Ablation study and complexity comparison on ImageNet-LT. ResNet-50 is used as backbone for training 90 epochs. PN: the number of parameters. GFLOPs: the number of floating-point operation of a forward pass for an input image.}
\begin{center}
\begin{tabular}{l |  c | c c | c| c c c }
\hline
\multirow{2}*{Method} & \multirow{2}*{PN} & \multicolumn{2}{c|}{GFlops} & \multirow{2}*{ALL} & \multirow{2}*{Many} &\multirow{2}*{Med.} &\multirow{2}*{Few} \\
\cline{3-4}
 & & Train &Test & & &  \\
\hline
cRT~\cite{Decoupling} &25.6M &4.100 &4.100 &47.3 &58.8 &44.0 &26.1\\
\hline
 RBMC-wo-R\&MC & 27.6M & 4.100 & 4.100 & 49.9 & 59.7 & 47.7 & 30.1 \\
RBMC-wo-R &28.1M &4.101 &4.101 &50.4 &59.9 &48.0 &31.6\\
RBMC-wo-MC  &27.6M &4.102 &4.102 &51.0 &62.9 &48.1 &27.3\\
RBMC &28.1M &4.103 &4.103 &\textbf{51.4} &63.3 &48.5 &28.5 \\
\hline
+FCM-$m$=1 &- &- &- &51.7 &61.1 &49.6 &32.8\\
+FCM &28.1M &4.113 &4.103 &\textbf{52.3} &61.2 &50.5 &33.6\\
\hline
+FCM-LCM-fix &- &- &- &52.6 &62.0 &50.5 &33.1\\
+FCM-LCM-SV &- &- &- &52.7 &61.8 &50.8 &33.4 \\
+FCM-LCM&28.1M & 4.115& 4.103& \textbf{53.2}&62.3 &51.2 &33.8\\
\hline
\end{tabular}
\end{center}
\label{Tab7}
\end{table}

 \begin{figure}[t]
\centering
\captionsetup{font={small}}
\includegraphics[width=1.0\linewidth]{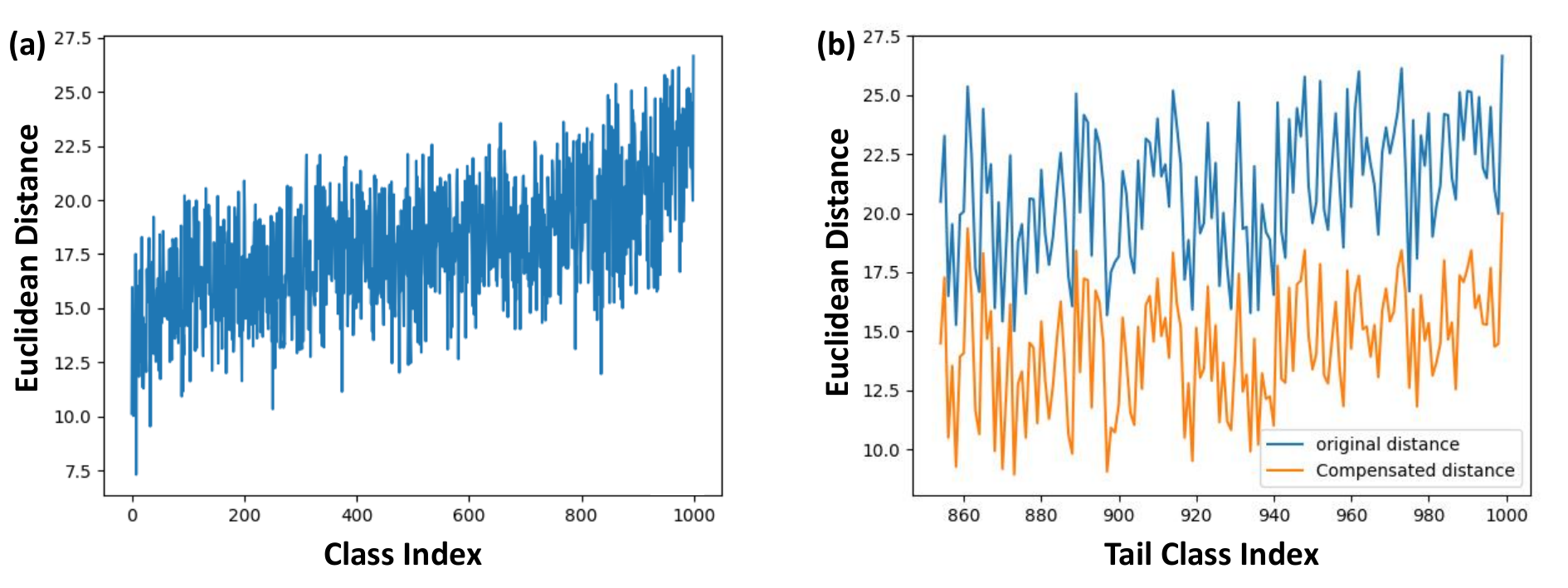}
\caption{(a) For each category, the average distance of all the test features to the closest original training feature of this category. (b) For each \textit{tail} category, the average distance of all the test features to the closest \textit{original training feature (blue)} / \textit{compensated feature by FCM (orange)} of this category.}
\label{fig7}
\end{figure}

\noindent
\textbf{Effectiveness of FCM.} \quad
To show the separate influence of FCM and LCM, we provide a variant of DCRNets that adds FCM alone (+FCM). As shown in Tab.~\ref{Tab7}, FCM brings remarkable improvements on few-shot split by up to $5.1\%$. The significant improvements indicate that it is important to compensate the feature drift between training and test data on tail classes. Fig.~\ref{fig7} (a) visualizes the average distance of all test features of each category to the closest training feature of this category. As shown, the distance goes large as the number of training instances decreases. The phenomenon reveals that the training and test features of a tail category do not occupy the same feature space. By contrast, as shown in Fig~\ref{fig7} (b), the compensated features of FCM are closer to the test features. As a consequence, FCM can make the classification boundary better adapt to the test data on tail classes.

We then investigate whether the \textit{multi-mode} feature drift compensation is necessary. We explore a variant \textbf{FCM-m=1} that estimates a \textit{uni-mode} feature drift direction for each tail category. It is equivalent to setting the number of selected head categories ($m$ in Eq.~\ref{eq1}) to 1. As shown in Tab.~\ref{Tab7}, FCM-m=1 leads to marginal improvements and FCM outperforms FCM-m=1 by $0.6\%$. The results verify the hypothesis that the features from one tail category could drift towards \textit{multiple} head categories. Notably, \cite{allen2019infinite} also  observe  the test data distribution of each few-shot category is clearly not uni-modal. So it is necessary for FCM to estimate a \textit{multi-mode} feature drift vector to better fit the complex test distribution of tail category. 
\\

\noindent
\textbf{Effectiveness of LCM.} \quad
Next, we assess the effectiveness of LCM. As shown in Tab.~\ref{Tab7}, LCM further achieves $0.9\%$ overall top-1 gain over FCM. Also, we observe that LCM improves the accuracy of head and tail classes simultaneously. The improvements can be attributed to the effectiveness of integrating uncertainty into the estimation of feature drift directions. The uncertainty of LCM can prevent the subsequent classifier from over-fitting  to FCM errors, thereby improving classifier's generalization. 
Then, we investigate the influence of \textit{Class Adaptive Translation Coefficient} ($\beta_t$ in Eq.~\ref{eq7}). We provide a variant \textbf{LCM-fix} that uses a fix translation coefficient for each class. LCM-fix is equivalent to setting $\beta_t$ to 1. As shown in Tab.~\ref{Tab7}, LCM-fix brings negligible gain over FCM, which indicates the necessity of using a larger translation coefficient for the more minority category. 

As discussed in Sec.~ \ref{Section3.2}, in order to deal with the smaller observed variance of a rarer class, previous work (\textbf{SV})~\cite{liu2020deep} uses a shared variance for all classes.  Tab.~\ref{Tab7} compares the proposed class adaptive strategy with SV. We observe that LCM outperforms \textbf{LCM-SV} by $0.5\%$. One possible explanation is that the unrelated classes, such as person and bag, generally have different intra-class variations. So for a specific tail class, the shared variance among all classes may lead to unreasonable translation directions, resulting in disturbed features. Differently, LCM uses the estimated variance from samples of the target class, which ensures the rationality of variation directions. In this way, the generated features still live in the true feature manifold, which can effectively benefit the learning of classifier. 
\\

\begin{table}[t]
\centering
\small
\caption{Combination FCM and LCM with various long-tailed methods on ImageNet-LT dataset. ResNet-50 is used as backbone trained for 90 epochs.}
\begin{center}
\begin{tabular}{ l | c c | c }
\hline
Method & FCM & LCM & ImageNet-LT\\
\hline
\multirow{3}*{BBN\cite{BBN}} & & &49.5\\
 & $\checkmark$ & &51.3 \\
  & $\checkmark$ & $\checkmark$ &\textbf{51.9} \\
\hline
\multirow{3}*{BALMS \cite{Balanced}} & & &50.7 \\
 &$\checkmark$ & & 52.3 \\
  & $\checkmark$ &$\checkmark$ &\textbf{53.0} \\
\hline
\multirow{3}*{LDAM \cite{cao2019learning}} & & &51.5 \\
 & $\checkmark$ & &52.5 \\
  & $\checkmark$ &$\checkmark$ &\textbf{53.3} \\
\hline
\end{tabular}
\end{center}
\label{Tab8}
\end{table}

\noindent
\textbf{Complexity Comparisons.}
To illustrate the cost of DCRNets, we report the number of parameters (PN) and the number of floating-point operations (GFLOPs). As shown in Tab.~\ref{Tab7}, RBMC only introduces $2.5$M parameters compared to a naive linear classifier (cRT). In addition, RBMC requires $4.103$ GFLOPs in a single forward pass for a $224\times 224$ input image, corresponding to only $0.07\%$ relative increase over cRT. FCM and LCM incurs $0.010$ and $0.002$ GFLOPs during training phase respectively. The increased cost of FCM mainly comes from  the extra classification of  generated compensated features. And the increased cost of LCM comes from the computation of logit compensation in Eq.~\ref{eq13}. Both operations can be worked out by matrix multiplication thus occupy little training time in GPU libraries. During test phase, FCM and LCM are discarded and DCRNets only introduces negligible inference time.
\\

\noindent
\textbf{Generalization of DCRNets across Different Backbones.} \quad
To investigate the effectiveness and generalization of DCRNets across different backbones, we also provide the ablation study based on ResNet-10 and ResNeXt-50 on ImageNet-LT dataset. As shown in Tab.~\ref{Tab6}, the proposed RBMC, FCM and LCM can consistently improve the results across different backbones. We also observe that: (1) the improvements of FCM and LCM are more obvious for deeper network. Specifically, the compensation modules, FCM and LCM, bring $1.0\%/1.8\%/1.9\%$ gain with ResNet-10/ResNet-50/ResNeXt-50 as backbone. We argue that the deeper network has a higher risk of overfitting on tail classes, resulting in a larger feature drift between training and test data. Thus the compensation modules could be more effective on deeper networks.  (2) The result of DCRNets based on ResNet-50  is much higher than the results of  baseline cRT~\cite{Decoupling} based on ResNeXt-50 (+$3.7\%$), where cRT based on ResNeXt-50 has more parameters and computational complexity. This comparison demonstrates that the improvement of DCRNets does not rely on the extra parameters and computational load.
\\

\noindent
\textbf{Generalization of DCRNets across Different benchmarks.} \quad
We further provide the ablation study on iNaturalist18. As shown in Tab.~\ref{Tab6}, the proposed modules, RBMC, FCM and LCM, consistently improve the performance across different benchmarks. Notably, FCM and LCM bring $2.6\%$ gain on iNaturalist18, which is higher than $1.8\%$ gain on ImageNet-LT. The reason is that iNaturalist18 contains a large number of \textbf{fine-grained} classes. The tail classes would be more similar to head classes of the same superclass, resulting in a more serious feature drift. So FCM and LCM, which compensate the feature drift, could be more effective on the fine-grained iNaturalist18 benchmark.

\begin{figure}[t]
\centering
\includegraphics[width=0.85\linewidth]{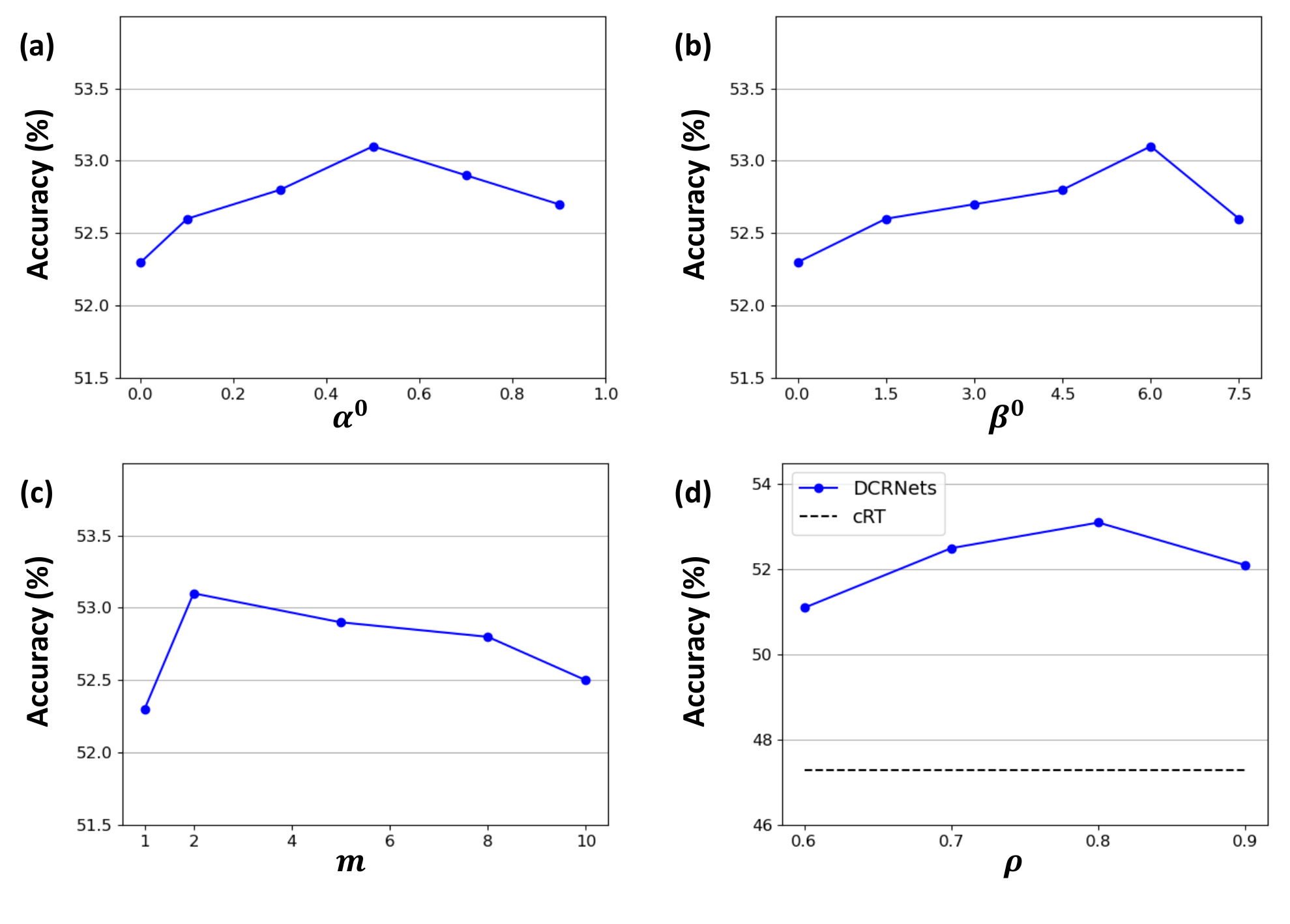}
\caption{Hyper-parameter analysis of DCRNets on ImageNet-LT. ResNet-50 is used as backbone for 90 epochs training. (a)  FCM compensation strength parameter $\alpha^0$. (b) LCM translation strength parameter $\beta^0$. (c) The number of selected head categories in FCM $m$. (d) Loss trade-off parameter $\phi$.}
\label{fig8}
\end{figure}

\subsection{Combination FCM/LCM with Various Methods}
Most studies on class imbalanced learning attempt to deal with \textbf{label shift} between training and test phase. Existing works generally assume that the extracted features of training and test data follow the same distribution. In this work, we point out that this assumption does not hold on tail classes. And we propose FCM and LCM to alleviate the feature drift of tail classes between training and test phase. To  demonstrate that addressing feature drift can consistently improve long-tailed performance, we combine FCM and LCM with several mainstream long-tailed methods, \textsl{i.e.}, BBN \cite{BBN}, BALMS \cite{Balanced} and LDAM \cite{cao2019learning}, based on their open-source codes. As shown in Tab.~\ref{Tab8}, FCM and LCM consistently bring about $2\%$ overall accuracy gains to various long-tailed methods, which validates the effectiveness and versatility of FCM and LCM. The consistent gains show that FCM and LCM can be used as general modules, which can combine with various long-tailed methods to enhance their feature generalization ability.

\subsection{Parameter Analysis}
\noindent \textbf{The affects of hyper-parameters.} \quad Fig.~\ref{fig8} explores the hyper-parameter sensitivity of our method. Hyper-parameter $\alpha^0$ and $\beta^0$ control the strength of feature drift compensation, $m$ is the number of selected head categories for compensation and $\phi$ balances the two losses in DCRNets. We respectively evaluate each hyper-parameter, where we change its value and fix other hyper-parameters to the optimal values. Notably, we select the hyper-parameters based on the performance of the validation set. And the phenomenon \textit{w.r.t} different hyper-parameters on validation set is consistent with that on test set. So we only present the performance on the test set in Fig.~\ref{fig8}.

Firstly, we change $\alpha^0$ from 0 to 1.0  and $\beta^0$ from 0 to 7.5. As shown in Fig.~\ref{fig8} (a) and  (b), the performances of different $\alpha^0>0/\beta^0>0$ consistently outperform that of  $\alpha^0=0/\beta^0=0$, which further verifies the effectiveness of FCM and LCM. In addition, the performances of different $\alpha^0>0$ and $\beta^0>0$ fluctuate in a small range ($\leq 0.5\%$), indicating the robustness of FCM and LCM to various $\alpha^0$ and $\beta^0$ in some degree. 

Secondly, we change $m$ from 1 to $10$. As shown in Fig.~\ref{fig8} (c), the performances of different $m > 1$ consistently outperform that of $m=1$, which implies the effectiveness of the \textit{multi-mode} drift vector in FCM. But there is an accuracy drop when $m$ is too big. The main reason is that the tail features could compensate towards unrelated head categories if the number of  selected head categories is too large. 

Thirdly, we change $\phi$ from 0.6 to 0.9. As shown in Fig.~\ref{fig8} (d), the performances of different $\phi$ consistently outperform cRT baseline, and our method achieves the best performance when $\phi=0.8$.  Notice that there will be $2\%$ accuracy drop when $\phi$ is too small ($\phi=0.6$). The main reason is that the feature extractor would over-emphasis on class balanced learning when $\phi$ is too mall, which damages the generalization of feature representations~\cite{Decoupling}. 

Overall, as shown in Fig.~\ref{fig8}, the performances of DCRNets fluctuates within $2\%$. The phenomenons show that our method is not sensitive to the hyper-parameters in some degree, and can achieve competitive results under a wide range of hyper-parameters values.

\noindent \textbf{The affects of batch sizes of two branches.} \quad
We then explore how  batch sizes of Uniform Branch and Class-Balanced Branch affect the performance. Denote the batch size of uniform branch and class-balanced branch as $n_1$ and $n_2$ respectively, Tab.~\ref{Tab9} shows the performance of DCRNets with different $n_1$ and $n_2$ on ImageNet-LT.  As shown, $n_1/n_2=192/64$ outperforms $n_1/n_2=128/128$ and $n_1/n_2=64/192$ by a large margin. We argue that using a too large $n_2$ would hurt feature representation. In particular, during each training iteration, a batch of $n_1$ examples and $n_2$ examples are sampled using uniform sampler and class-balanced sampler respectively. 
So using a too large $n_2$  would increase the risks of over-fitting the tail data (by over-sampling) and under-fitting the head data (by under-sampling). This could unexpectedly damage the representative ability of the learned deep features \cite{Decoupling}. In addition, due to the small number of parameters in the classifier, a small $n_2$ is usually sufficient to learn a good balanced classifier. Therefore, it is reasonable to set $n_1>n_2$. Tab.~\ref{Tab9} also shows that $n_1/n_2=192/64$ achieves marginally better result than $n_1/n_2=224/32$. We argue that using a \textit{too small} $n_2$ is inadequate to learn a class-balanced classifier. Therefore, we set $n_2=n_1/3$ on all benchmarks, which is capable of simultaneously learning a good feature representation and class-balanced classifier.

\begin{table}[t]
\centering
\small
\caption{ The performance of DCRNets with different batch sizes for uniform branch and class-balanced branch. ResNet-50 is used as backbone for training 90 epochs on ImageNet-LT. $n_1/n_2$ denote the batch size of uniform/class-balanced branch.}
\vspace{-3.5mm}
\begin{center}
\begin{tabular}{l | c c c c}
\hline
$n_1/n_2$ & $224/32$ & $192/64$ & $128/128$ & $64/192$ \\
\hline
DCRNets &52.9 &\textbf{53.2} &52.3 & 47.6\\
\hline
\end{tabular}
\end{center}
\label{Tab9}
\end{table}

\subsection{Evaluation on Class Incremental Learning}
For Class Incremental Learning (CIncL), an underlying problem is that the ratio of the number of new samples to that of old samples (preserved exemplars) could be very high, resulting in severe class imbalance. Thus DCRNets that address class imbalance can be leveraged for CIncL. In this section, we additionally evaluate DCRNets on CIncL.

\textbf{Datasets and Implementations.} \quad
We conduct CIncL experiments on CIFAR100 \cite{krizhevsky2009learning} following a closely related work \cite{zhao2020maintaining}. CIFAR100 contains 60,000 samples of $32\times 32$ images for 100 classes. 
Following the exact setting in \cite{zhao2020maintaining}, we deploy a 32-layer ResNet \cite{residual} as the baseline architecture for CIFAR100. We use SGD to train our model and set the batch size for uniform branch and class balanced branch to 128 and 32 respectively. The learning rate starts from 0.1 and reduces to 0.1 of the previous learning rate after 120, 180, 240 epochs (250 epochs in total). The hyper-parameters $m$, $\alpha^0$ and $\beta^0$ are set to 2, 0.5 and 0.5 respectively. The weighting factor of loss function  $\phi$ is set to 0.9.

\textbf{Memory Budge.} \quad
We follow the same data reply setting used in \cite{zhao2020maintaining}. We store  2,000 samples for old classes, and select rehearsal exemplars based on herding selection \cite{welling2009herding}. More classes have been seen, fewer images can be retained per class. As a result, the problem of class imbalance becomes more serious.

\textbf{Benchmark Protocol.} \quad
We follow the common protocol used in \cite{zhao2020maintaining}. On CIFAR100 benchmark, 100 classes are evenly split into 2/5/10 incremental steps. In each step, the model is evaluated on the test data for all seen classes. Since the first step is not related to class incremental learning actually, we report the average top-1 accuracy over all incremental phases except the first step. 

\begin{table}[t]
\centering
\small
\caption{Class incremental learning performance (top-1 accuracy $\%$) on CIFAR100 with 2, 5 and 10 incremental steps. The average results over all incremental steps except the first step are reported. \textbf{CIncL} stands for Class Incremental Learning methods and \textbf{CImbL} indicates Class Imbalanced Learning inspired methods. We produce the results using their public code. }
\begin{center}
\begin{tabular}{ l|l | c  | c | c  }
\hline
  &\multirow{2}*{ Method} & \multicolumn{3}{c}{$\#$incremental steps} \\
 \cline{3-5} 
 & & 2&5&10\\
\hline
\multirow{5}*{CIncL} 
&LwF \cite{li2017learning} & 52.6 & 47.1 & 39.7 \\
&iCaRL \cite{rebuffi2017icarl} &62.0 &63.3 & 61.6\\ 
&EEIL \cite{castro2018end} &60.8 & 63.7 & 63.6 \\
&IL2M \cite{belouadah2019il2m} &65.0 &65.5 &63.7 \\
&BiC \cite{wu2019large} &65.3 & 66.0 &63.7  \\
&MDFCIL \cite{zhao2020maintaining} &66.5 & 67.2 & 64.7 \\
\hline
\multirow{5}*{CImbL}
&$\tau$-norm \cite{Decoupling} &65.7 &66.5 &64.4 \\
&cRT \cite{Decoupling} &66.2 &66.8 &65.5 \\
&LWS \cite{Decoupling} &66.6 & 66.5 &63.5\\
&LADE \cite{hong2020disentangling} &66.6 &67.4 &64.4\\
\cline{2-5} 
&\textbf{DCRNets} &\textbf{68.2} &\textbf{68.5} &\textbf{67.4}\\ 
\hline
\end{tabular}
\end{center}
\label{Tab10}
\end{table}

\textbf{Comparison to Other Methods.} \quad
We compare DCRNets with several competitive or representative methods, including Class Incremental Learning methods (CIncL) and Class Imbalanced Learning inspired methods (CImbL). As for CIncL, we compare LwF \cite{li2017learning}, iCaRL \cite{rebuffi2017icarl}, EEIL \cite{castro2018end}, IL2M \cite{belouadah2019il2m}, BiC \cite{wu2019large} and MDFCIL \cite{zhao2020maintaining}. 
Note that \textbf{EEIL, BiC,  IL2M and MDFCIL attempt to address the class imbalance problem in CIncL}. As for CImbL-inspired methods, we compare cRT \cite{Decoupling}, $\tau$-norm \cite{Decoupling}, LWS \cite{Decoupling} and LADE \cite{hong2020disentangling}. We produce the results using their public codes.
The compared results are shown in Tab.~\ref{Tab10}.

As for CIncL methods, LwF \cite{li2017learning}  achieves much lower performance than others. It is reasonable since it incorporates no imbalance technique, making it vulnerable to class imbalance. EEIL \cite{castro2018end}, BiC  \cite{wu2019large},  IL2M \cite{belouadah2019il2m} and MDFCIL \cite{zhao2020maintaining} propose different bias correction techniques to boost the  performance. MDFCIL \cite{zhao2020maintaining} is the most effective one. In MDFCIL, by normalizing the classifier weights to have similar norms, the bias induced by different numbers of samples can be removed. As for CImbL methods, cRT \cite{Decoupling}, $\tau$-norm \cite{Decoupling}, LWS \cite{Decoupling} and LADE \cite{hong2020disentangling} add a class-balanced step to correct the outputs, which achieve similar performance with MDFCIL. Our DCRNets consistently outperform the compared methods. Specifically, the overall performance on 10 incremental steps is improved by $2.7\%$ accuracy compared to MDFCIL, and $1.9\%$ compared to cRT. The reason is that all above methods assume that the extracted features of training and test data follow a same distribution, which is not hold on imbalanced data of CIncL. Our method effectively reduces the unfavorable feature drift between training and test phases to better address imbalance issue, thereby benefiting  incremental learning. 

Overall, Tab.~\ref{Tab10} indicates that the techniques in CImbL can be applied in CIncL as well. And our approach is more effective to handle class imbalance in CIncL. Our DCRNets can achieve better 
results compared to state-of-the-art approaches under different incremental settings.

\section{Conclusion}
We propose a framework, DCRNets, for class imbalanced learning. Firstly, we observe that there is a severe feature drift between training and test data, especially on tail class. So we design FCM and LCM to estimate the feature drift and compensate for it. Our study suggests that feature compensation is important to achieve a generalizable classifier on imbalanced data.  Secondly, we observe that CBS essentially leads to underfitting on head classes. So RBMC is designed to add a residual path from Uniform Branch to facilitate classifier learning. Our study suggests that uniform learning is also useful for learning a class-balanced classifier. 
Our overall framework is \textit{generic} and can be easily incorporated into existing Class Imbalanced methods to boost their performance. We also evaluate our approach on Class Incremental Learning to show its universality.

\section{Further Work}
In this part, we briefly discuss the feasibility of our approach on other tasks. We leave the specific application of our proposed method to other tasks as a further work.

\noindent \textbf{\underline{Vehicle/Person ReID}}. \quad  This work mainly explores deep classification task. We argue our method can be  also used to Vehicle/Person reID (\textit{metric learning} task) when the identities present an imbalanced distribution. 

\textbf{Firstly, the overall two-branch framework is applicable for reID}. The recently most successful approaches~\cite{sun2020circle} show that the metric loss~\cite{hermans2017defense,sun2020circle} is superior to reID.  However, on imbalanced setting, metric loss would concentrate more on head identities than tail ones since the comparison chances for tail identities reduce quadratically. So the metric training process is dominated by head identities, which could distort overall feature space~\cite{liu2020deep}. In our framework, the metric loss can be added on the class-balanced branch to perform balanced metric learning, to learn better feature representation for reID. 

\textbf{Secondly, our FCM and LCM which compensate tail identities with higher diversity are applicable for reID.} As pointed by~\cite{liu2020deep}, on reID with imbalanced data, the head identity often occupies a large spatial span, while the tail identity often occupies a very small spatial span on feature space due to lack of intra-class diversity. This uneven distribution distorts overall feature space, consequentially compromising the discriminative ability of learned features. In our framework, FCM translates the tail features along meaningful nearest-neighbor directions, and LCM augments tail features along intra-class variations more strongly. Thus FCM and LCM can provide a larger space span with higher diversity for tail identities, thereby alleviating the distortion of feature space and improving representation learning on reID with imbalanced identities.

\noindent \textbf{\underline{Unsupervised Learning}}. \quad
\textit{Unsupervised Clustering Algorithm}~\cite{caron2018deep,amrani2022self} is an important line of unsupervised learning. Unsupervised Clustering usually alternately learns the network parameters and the cluster assignments (pseudo-labels) for unlabeled data. We argue that our method could be applied to unsupervised clustering methods in the following two ways:
\textbf{(1) Our two-branch framework could compensate for imbalanced clusters.} ~\cite{caron2018deep}  observes that clustering algorithm, \textit{e.g.}, k-means, is vulnerable to imbalanced clusters (pseudo-labels), where vast majority of examples  are assigned to a few clusters. For unsupervised learning, our two-branch framework could use \textit{pseudo-labels} to perform uniform and class-balanced sampling. In this way,  the class-balanced branch could compensate for imbalanced clusters. \textbf{(2) Our FCM and LCM could reduce clustering errors.} ~\cite{cho2022part,ge2020self}  observe that the clustering pseudo-labels are inherently noisy, \textit{e.g.}, the samples sharing the same true class could be split  into two or more clusters. The key of FCM and LCM is to synthesize new samples along the directions to its nearest neighbors. Therefore, FCM and LCM can be used to synthesize intermediate samples between two neighbor sub-clusters with the same true class. The synthesized samples can help correctly merge the two sub-clusters into one cluster. Therefore, we argue that FCM and LCM are able to reduce clustering errors and thus improve unsupervised learning.

\noindent \textbf{\underline{Unsupervised Domain Adaptation (UDA)}}. \quad 
The key problem of UDA is the feature drift between source-domain and target-domain data. Our FCM and LCM are designed to deal with feature drift between training and test data, which can also address the feature drift issue on UDA. That is, FCM and LCM can translate the source features towards the target domain, by setting the feature drift vector of class $c$ ($\boldsymbol{\delta_c}$) as the distance of class-$c$ prototype between source data and target data. By adding the feature drift vector to original source features, the new generated source features will be closer to the target domain, which is  conducive to train a more transferable classifier.

\noindent \textbf{\underline{Continual UDA.}} \quad
The  work ~\cite{taufique2021conda} proposed Continual UDA where unlabeled target-domain samples are received in small batches and adaptation is performed continually. The target-domain data is collected \textit{randomly} and \textit{incrementally} in \textit{small} batches, which easily causes the observed data to be class-imbalanced, especially in the early continual learning stage. Therefore, our framework can also be applied to Continual UDA to address its class imbalance issue.


%



\ifCLASSOPTIONcompsoc
  \section*{Acknowledgments}
\else
  \section*{Acknowledgment}
\fi

This work is partially supported by National Key R\&D Program of China (No. 2018AAA0102402) and Natural Science Foundation of China (NSFC): 61976203, 62276246 and U19B2036, and
in part by the National Postdoctoral Program for Innovative Talents under Grant BX20220310.

\bibliographystyle{ieeetr}
\bibliography{egbib}

\ifCLASSOPTIONcaptionsoff
  \newpage
\fi

\begin{IEEEbiography}[{\includegraphics[width=1in,height=1.25in,clip,keepaspectratio]{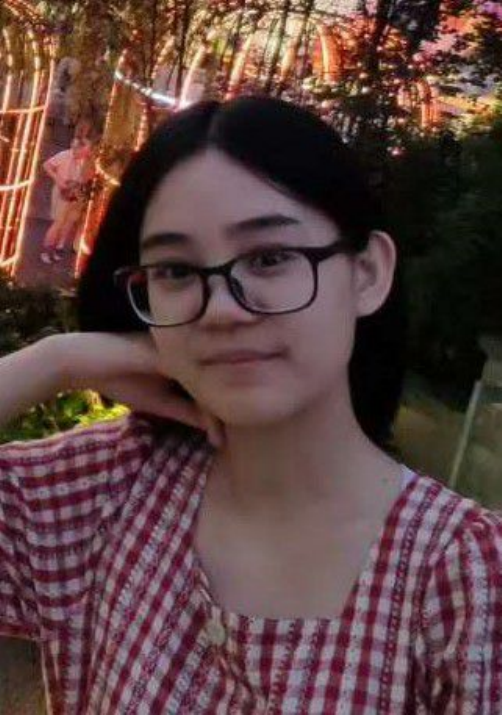}}]{Ruibing Hou}
received the BS degree in Northwestern Polytechnical University, Xi’an, China, in 2016. She received PhD degree in computer science from the Institute of Computing Technology, Chinese Academy of Sciences, Beijing, China, in 2022. She is currently a post-doctorial researcher with the Institute of Computing Technology, Chinese Academy of Sciences. Her research interests are in machine learning and computer vision. She specially focuses on person re-identification, long-tailed learning and few-shot learning. 
\end{IEEEbiography}

\begin{IEEEbiography}[{\includegraphics[width=1in,height=1.25in,clip,keepaspectratio]{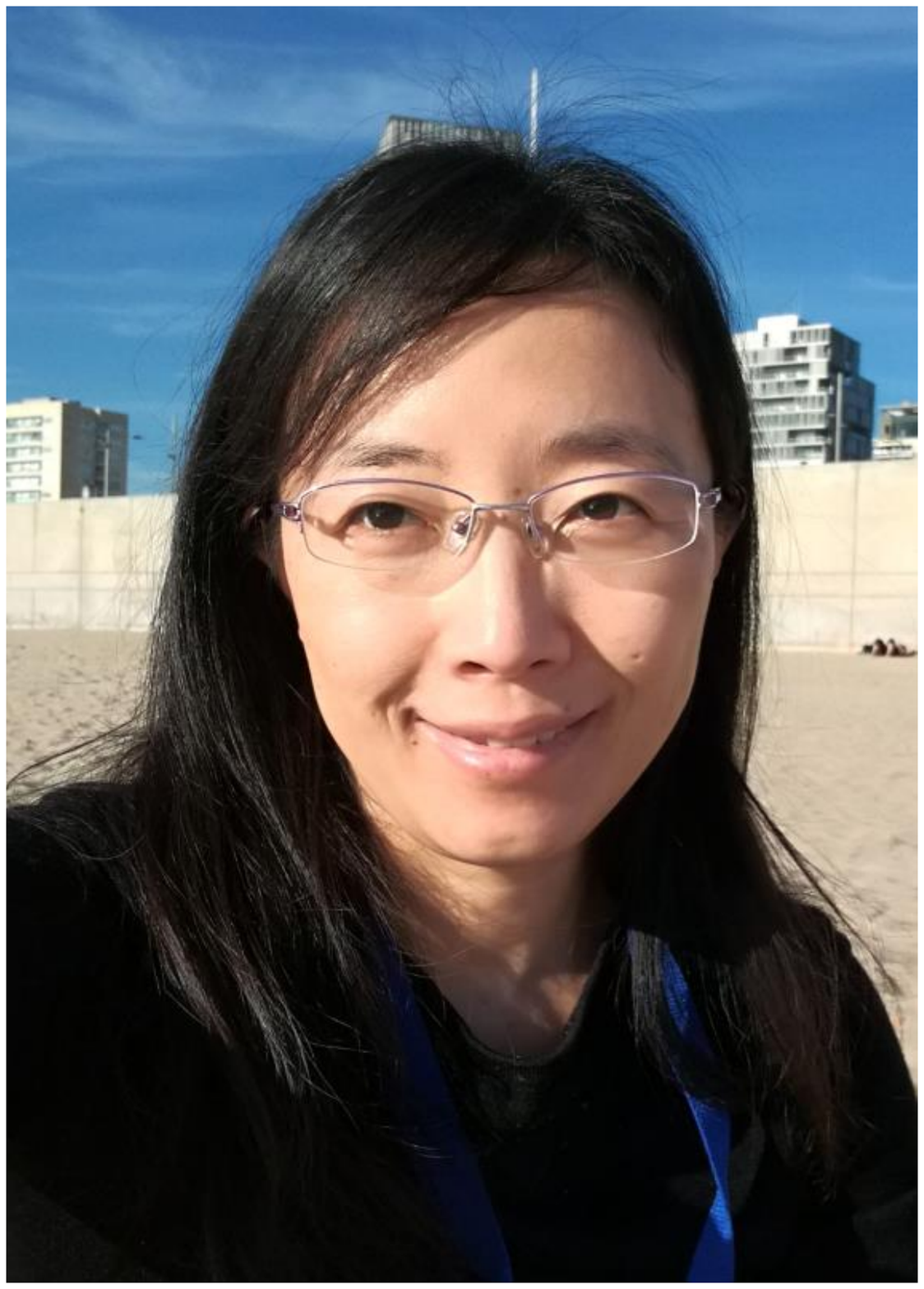}}]{Hong Chang}
received the Bachelor’s degree from Hebei University of Technology, Tianjin,
China, in 1998; the M.S. degree from Tianjin University, Tianjin, in 2001; and the Ph.D. degree from Hong Kong University of Science and Technology, Kowloon, Hong Kong, in 2006, all in computer science. She was a Research Scientist with Xerox Research Centre Europe. She is currently a Researcher with the Institute of Computing Technology, Chinese Academy of Sciences, Beijing, China. Her main research interests include algorithms and models in machine learning, and their applications in pattern recognition and computer vision.
\end{IEEEbiography}

\begin{IEEEbiography}[{\includegraphics[width=1in,height=1.25in,clip,keepaspectratio]{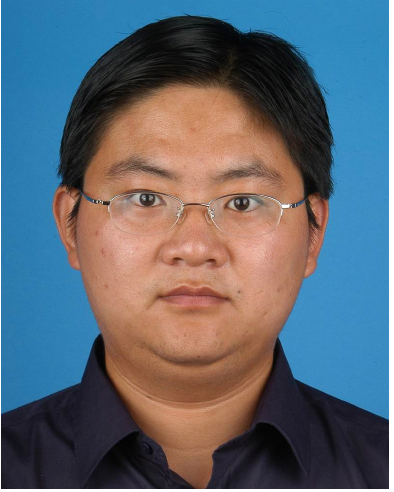}}]{Bingpeng Ma}
received the BS degree in mechanics, in 1998 and the MS degree in mathematics, in 2003 from the Huazhong University of Science and Technology, respectively. He received the PhD degree in computer science from the Institute of Computing Technology, Chinese Academy of Sciences, P.R. China, in 2009. He was a post-doctorial researcher with the University of Caen, France, from 2011 to 2012. He joined the School of Computer Science and Technology, University of Chinese Academy of Sciences, Beijing, in March 2013 and now he is a professor. His research interests cover computer vision, pattern recognition, and machine learning. He especially focuses on person re-identification, face recognition, and the related research topics.
\end{IEEEbiography}

\begin{IEEEbiography}[{\includegraphics[width=1in,height=1.25in,clip,keepaspectratio]{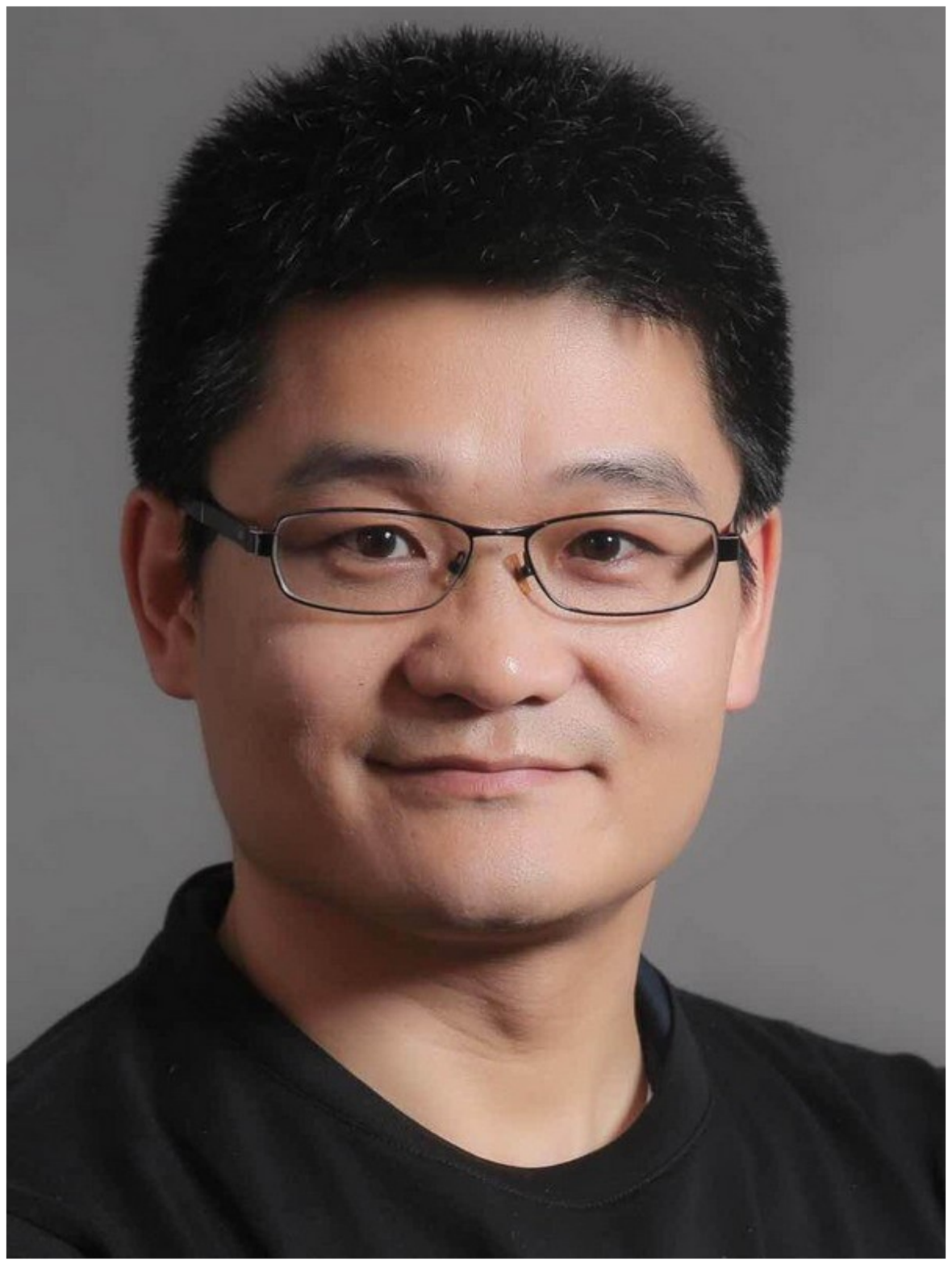}}]{Shiguang Shan}
(M’04-SM’15) received Ph.D. degree in computer science from the Institute of Computing Technology (ICT), Chinese Academy of Sciences (CAS), Beijing, China, in 2004. He has been a full Professor of this institute since 2010 and now the deputy director of CAS Key Lab of Intelligent Information Processing. His research interests cover computer vision, pattern recognition, and machine learning. He has published more than 300 papers, with totally more than 20,000 Google scholar citations. He served as Area Chairs for many international conferences including CVPR, ICCV, AAAI, IJCAI, ACCV, ICPR, FG, etc. And he was/is Associate Editors of several journals including IEEE T-IP, Neurocomputing, CVIU, and PRL. He was a recipient of the China’s State Natural Science Award in 2015, and the China’s State S\&T Progress Award in 2005 for his research work.
\end{IEEEbiography}

\begin{IEEEbiography}[{\includegraphics[width=1in,height=1.25in,clip,keepaspectratio]{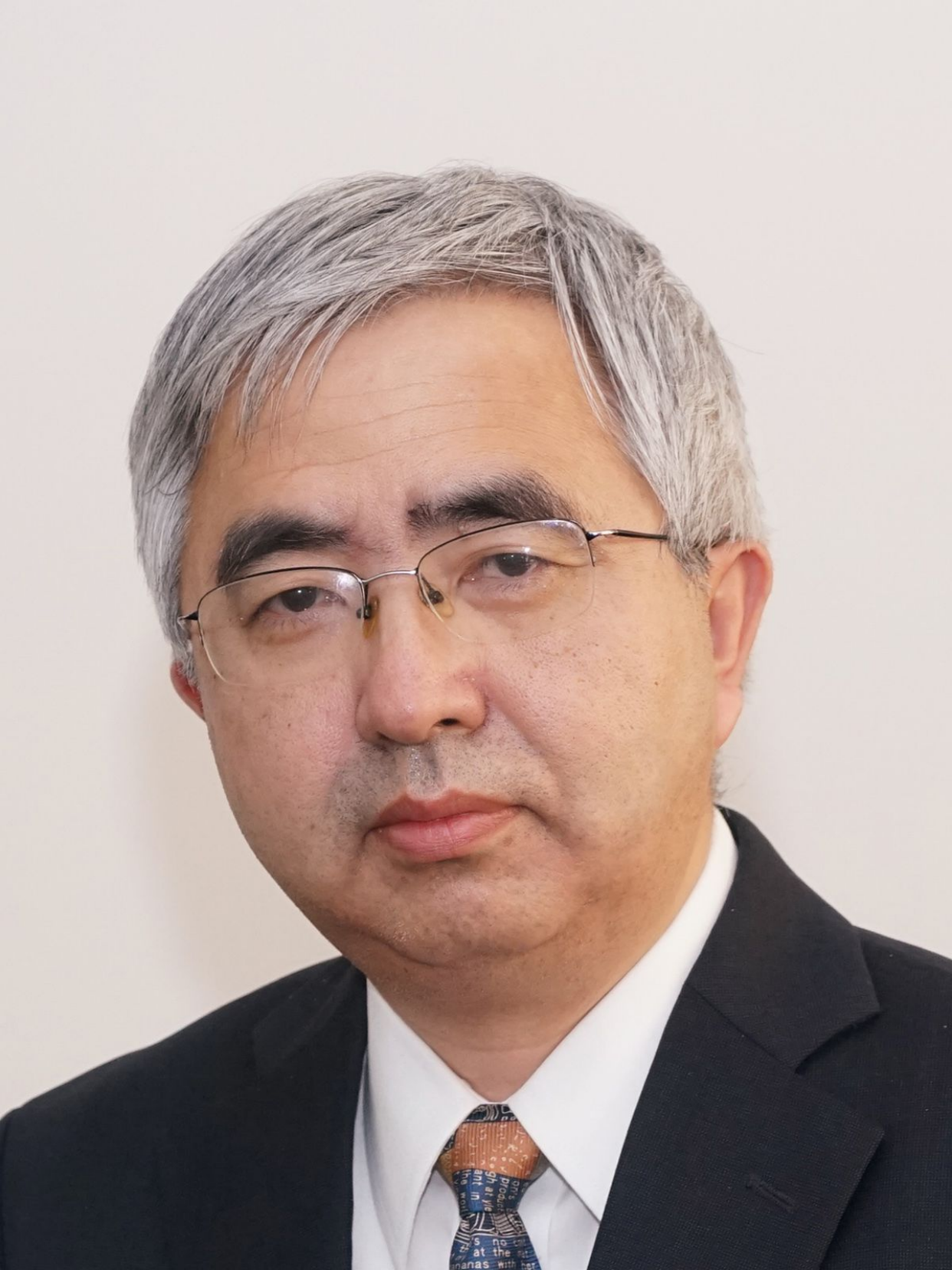}}]{Xilin Chen}
is a professor with the Institute of Computing Technology, Chinese Academy of Sciences (CAS). He has authored one book and more than 400 papers in refereed journals and proceedings in the
areas of computer vision, pattern recognition, image processing, and multimodal interfaces. He is currently an information sciences editorial board member of Fundamental Research, an editorial board member of Research, a senior editor of the Journal of Visual Communication and Image Representation, and an associate editor-in-chief of the Chinese Journal of Computers, and Chinese Journal of Pattern Recognition and Artificial Intelligence. He served as an organizing committee member for multiple conferences, including general co-chair of FG 2013 / FG 2018, VCIP 2022, program co-chair of ICMI 2010 / FG 024, and an Area Chair / Senior PC of ICCV / CVPR / ECCV / ICMI for more than 10 times. He is a fellow of the ACM, IEEE, IAPR, and CCF.
\end{IEEEbiography}




\end{document}